\documentclass[10pt,twocolumn,letterpaper]{article}

\usepackage{iccv}
\usepackage{times}
\usepackage{epsfig}
\usepackage{graphicx}
\usepackage{amsmath}
\usepackage{amssymb}

\usepackage{microtype}
\usepackage{graphicx}
\usepackage{subfigure}
\usepackage{booktabs} % for professional tables

\usepackage{url}

\usepackage{nicefrac}       % compact symbols for 1/2, etc.
\usepackage{makecell}

\usepackage{algorithm}
\usepackage{algpseudocode}
\usepackage{relsize}

\usepackage{amsmath,amssymb,amsfonts,amsthm,mathtools,bm}
\usepackage{enumitem}
\usepackage{multirow}
\usepackage{bm}
\usepackage{array}
\usepackage{color}
\usepackage{xcolor}
\usepackage{wrapfig}
\usepackage{multicol}
\usepackage{diagbox}
\usepackage[accsupp]{axessibility}  % Improves PDF readability for those with disabilities.

\newtheorem{prop}{Proposition}

\definecolor{customred}{RGB}{188, 50, 44}

\DeclareMathOperator*{\argmax}{argmax}
\definecolor{customred}{RGB}{192, 0, 0}
\newcommand*\Let[2]{\State #1 $\gets$ #2}

\usepackage[pagebackref=true,breaklinks=true,letterpaper=true,colorlinks,bookmarks=false]{hyperref}

\iccvfinalcopy % *** Uncomment this line for the final submission

\def\iccvPaperID{7526} % *** Enter the ICCV Paper ID here

\ificcvfinal\fi

\begin{document}

%%%%%%%%% TITLE
\title{Cluster-Promoting Quantization with Bit-Drop for Minimizing Network Quantization Loss}

\author{Jung Hyun Lee$^{1}$\thanks{Equal contribution},  Jihun Yun$^{1}$\footnotemark[1], Sung Ju Hwang$^{1,2}$, Eunho Yang$^{1,2}$\\
$^{1}$Korea Advanced Institute of Science and Technology (KAIST), $^{2}$AITRICS \\
{\tt\small \{onliwad101, arcprime, sjhwang, eunhoy\}@kaist.ac.kr}
}

% \author{
% 	Jung Hyun Lee$^{1}$\thanks{Equal contribution}, Jihun Yun$^{1}$\footnotemark[1], Sung Ju Hwang$^{1,2}$, Eunho Yang$^{1,2}$\\
% 	{\small onliwad101@kaist.ac.kr, arcprime@kaist.ac.kr, sjhwang@kaist.ac.kr eunhoy@kaist.ac.kr}\\
% 	{\small $^{1}$Korea Advanced Institute of Science and Technology (KAIST), $^{2}$AITRICS}%\\ 
% 	%{\small $^{2}$AITRICS, Seoul, South Korea}
% } 

\maketitle
\ificcvfinal\thispagestyle{empty}\fi

\begin{abstract}
    Network quantization, which aims to reduce the bit-lengths of the network weights and activations, has emerged for their deployments to resource-limited devices. 
    Although recent studies have successfully discretized a full-precision network, they still incur large quantization errors after training, thus giving rise to a significant performance gap between a full-precision network and its quantized counterpart. In this work, we propose a novel quantization method for neural networks, \emph{Cluster-Promoting Quantization (CPQ)} that finds the optimal quantization grids while naturally encouraging the underlying full-precision weights to gather around those quantization grids cohesively during training. This property of CPQ is thanks to our two main ingredients that enable differentiable quantization: i) the use of the categorical distribution designed by a specific probabilistic parametrization in the forward pass and ii) our proposed multi-class straight-through estimator (STE) in the backward pass. Since our second component, multi-class STE, is intrinsically biased, we additionally propose a new bit-drop technique, \emph{DropBits}, that revises the standard dropout regularization to randomly drop bits instead of neurons.
    As a natural extension of DropBits, we further introduce the way of learning heterogeneous quantization levels to find proper bit-length for each layer by imposing an additional regularization on DropBits. We experimentally validate our method on various benchmark datasets and network architectures, and also support a new hypothesis for quantization: learning heterogeneous quantization levels outperforms the case using the same but fixed quantization levels from scratch.  
\end{abstract}
\vspace{-0.1cm}
\section{Introduction}\label{sec:intro}
Deep neural networks have achieved great success in various computer vision applications. However, the state-of-the-art neural network architectures including ResNet \cite{he2016deep} generally require too much computation and memory to be deployed to resource-limited devices. Therefore, researchers have explored diverse approaches to compress them to reduce memory usage and computation cost. 

% In this paper, we focus on \emph{neural network quantization}, which 
Among them, network quantization aims to reduce the bit-width of network parameters while maintaining competitive performance of a full-precision counterpart. One of the simplest methods is to round a weight or an activation of a network $x$ to $\widehat x = \alpha \lfloor \frac{x}{\alpha} + \frac{1}{2} \rfloor$ where $\alpha$ controls the grid interval size. However, this na\"ive approach incurs severe performance degradation mainly due to the \emph{quantization loss}. Given that if the underlying full-precision weights $x$ are clustered well around the optimal quantization grids, the performance difference between before and after the quantization can be marginal so that the performance of full-precision network can be preserved even with the quantized parameters. Hence, we focus on jointly finding the optimal quantization grids and clustering the underlying full-precision weights $x$ around those quantization grids cohesively. % to achieve comparable performance to a full-precision network. 

%Although the recent network quantization methods such as Variational Network Quantization (VNQ) \cite{Achterhold2018vnq} and Relaxed Quantization (RQ) \cite{louizos2018relaxed} take into account clustering while finding the optimal quantization grids, their clustering abilities are still suboptimal.
Some recent studies in fact have experimentally confirmed that their methods can partially give a clustering effect in the quantization process. VNQ \cite{Achterhold2018vnq} clusters the underlying full-precision weights $x$ around quantization grids using multi-spike-and-slab prior, but it is restricted only to ternary precision. RQ \cite{louizos2018relaxed} experimentally shows some clustering effects around several modes in low bit-width, but it does not equip any algorithm that explicitly encourages clustering around quantization grids. As a result, both methods incur a considerable performance gap between a full-precision network and its quantized counterpart. %  both VNQ and RQ strive to force network parameters to be concentrated with high density; however, 

In order to preserve the performance of a full-precision network in low bit-width, we propose the \emph{Cluster-Promoting Quantization (CPQ)} that not only finds the optimal quantization grids but also encourages the underlying full-precision weights $x$ to gather around those quantization grids cohesively in low bit-length regimes. Although CPQ does not have any explicit regularization or loss for clustering, the combination of the following two key components results in better clustering effect (and thus final performance) both theoretically and experimentally: i) choosing the mode of the categorical distribution parametrized by a particular probabilistic approach in the forward pass and ii) taking advantage of our multi-class straight-through estimator (STE) in the backward pass.

%CPQ theoretically and empirically show the superiority of clustering without any regularization or loss for clustering.

% Without any regularization or loss for clustering, we theoretically and empirically show that CPQ can help network parameters congregate around quantization grids cohesively in low bit-width regimes just by choosing the mode of the original categorical distribution parametrized by a particular probabilistic approach in the forward pass and taking advantage of our multi-class straight-through estimator (STE) in the backward pass that allows gradient-based optimization as well.

As our multi-class STE is biased like the original STE for the binary case \cite{Bengio2013ste}, we present a novel bit-drop technique named \emph{DropBits} to reduce the bias of the multi-class STE in CPQ. Motivated from Dropout \cite{srivastava2014dropout}, DropBits drops bits rather than neurons/filters to train low-bit neural networks under CPQ framework.

In addition, DropBits allows \emph{heterogeneous quantization}, which learns different bit-width per parameter/channel/layer by dropping redundant bits. DropBits with learnable bit-drop rates adaptively finds out the optimal bit-width for each group of parameters, possibly further reducing the overall bits. In contrast to recent studies \cite{wang2019haq,uhlich2020mixed} in heterogeneous quantization that exhibit almost all layers have \emph{at least} 4-bit, up to 10-bit, our method yields much more resource-efficient low-bit neural networks with \emph{at most} 4 bits for all layers.

With trainable bit-widths, we also articulate a \emph{new hypothesis for quantization} where one can find the learned % reduced
bit-width network (termed a `quantized sub-network') which can perform better than the network with the same but fixed bit-widths from scratch.

Our contribution is threefold:
% \vspace{-0.2cm}
\begin{itemize}[leftmargin=8mm]
	\item We propose a new quantization method, \textbf{C}luster-\textbf{P}romoting \textbf{Q}uantization (CPQ) that not only finds the optimal quantization grids but also encourages the underlying full-precision weights %network parameters 
	to congregate around those quantization grids cohesively in low bit-width regimes by the combination of a particular probabilistic parametrization for discretization and our multi-class straight-through estimator. We further present a novel bit-drop technique coined \textbf{DropBits} to reduce the bias of the multi-class straight-through estimator in CPQ.
% 	\vspace{-0.2cm}
	\item Extending DropBits technique, we propose a more resource-efficient heterogeneous quantization algorithm to curtail redundant bit-widths across groups of weights and/or activations (e.g. across layers) and verify that our method is able to find out `quantized sub-networks'.  
% 	\vspace{-0.2cm}
	\item We conduct extensive experiments on several benchmark datasets to demonstrate the effectiveness of our method. We accomplish new \textbf{state-of-the-art} results for ResNet-18 and MobileNetV2 on the ImageNet dataset when \emph{all} layers are uniformly quantized. 
\end{itemize}
% \vspace{-0.1cm}

\section{Related Work}\label{sec:related}

BinaryConnect \cite{Courbariaux2015binaryconnect} first attempted to binarize weights to $\pm 1$ by employing deterministic or stochastic operation. To obtain better performance, various studies \cite{Rastegari2016xnornet, Li2016ternary, Achterhold2018vnq, Shayer2018lrnet} have been conducted in binarization and ternarization.
% Another work on binary networks, Binary-Weight-Network \cite{Rastegari2016xnornet}, also binarizes weights by minimizing %the distance from weights to scaled binary matrix. In other words, they minimize 
% the distance $\|W - \alpha B\|_2$ where $\alpha \in \mathbb{R}_{++}$ manipulates the grid interval length and $B$ represents the binary weights. % and can be computed in a closed form.
% Enhancing the expressiveness of a model while keeping low-bit precision, TWN \cite{Li2016ternary} ternarizes weights to $\{0, \pm 1\}$ and update a scaling factor $\alpha$ controlling the grid size during training. VNQ \cite{Achterhold2018vnq} also ternarizes weights by introducing a specific multi-modal quantizing prior in a variational framework. Similarly, \cite{Shayer2018lrnet} proposed a discrete version of reparameterization trick \cite{kingma2015variational} for learning discrete-weight networks. 
Although these works effectively decrease the model size and raise the accuracy, they are limited to quantizing weights with activations remaining in full-precision. %reducing bit-width because they only quantize weights. 
To take full advantage of quantization at run-time, %of bit convolution kernels in quantized neural networks, 
it is necessary to quantize activations as well. 

% Recent algorithms
Researchers have recently focused more on simultaneously quantizing both weights and activations \cite{Zhou2016dorefa, yin2018bcgd, choi2018pact, zhang2018lqnets, gong2019dsq,jung2019learning, esser2020learned}. %To exploit efficiency of XNOR and bit-counting operations, XNOR-Networks \cite{Rastegari2016xnornet} binarize activations by bringing another scaling factor, which is also expressed as a closed-form solution. 
% The main example in this line of work is 
XNOR-Net \cite{Rastegari2016xnornet} %, the beginning of this line of work, 
exploits the efficiency of XNOR and bit-counting operations. QIL \cite{jung2019learning} also quantizes weights and activations by introducing parametrized learnable quantizers that can be trained jointly with weight parameters. 
% For more efficient training, DoReFa-Net \cite{Zhou2016dorefa} quantizes gradients as well to cut down time spent in backpropagation. 
\cite{esser2020learned} recently presented a simple technique to approximate the gradients with respect to the grid interval size to improve QIL.
Nevertheless, these methods do not quantize the first or last layer, which leaves a room to improve power-efficiency.

For ease of deployment in practice, it is inevitable to quantize weights and activations of all layers, which is the most challenging. 
% BNN \cite{Hubara2016bnn} quantizes all layers by extending the same technique as BinaryConnect \cite{Courbariaux2015binaryconnect} to activations. They proposed two binarization schemes: deterministic sign function binarization and stochastic binarization with the input-dependent probability. Although stochastic version seems advantageous in terms of performance like stochastic rounding \cite{gupta2015deep}, they employed the deterministic one due to hardware issues.
\cite{Achterhold2018vnq} proposed multi-spike-and-slab prior to allow multiple modes at quantization grids, but it is limited to ternary precision.
\cite{louizos2018relaxed} proposed to use the Gumbel-Softmax trick \cite{Jang2016gumbel,Maddison2016concrete}, but it does not cluster weights around quantization grids well.
\cite{jain2019tqt} presented efficient fixed-point implementations by formulating the grid interval size to the power of two, but they quantized the first and last layer to at least 8-bit. \cite{zhao2020linear} proposed to quantize the grid interval size and network parameters in batch normalization for the deployment of quantized models on low-bit integer hardware, but it requires a specific accelerator only for this approach. 

%Note that all the aforementioned quantization methods are concerned with fixed bit-width precision. 
% \vspace{-0.1cm}
As another line of work, \cite{fromm2018heterogeneous} proposed a heterogeneous binarization given pre-defined bit-distribution. 
% but their parameter-wise mixed precision brings about the limited applicability. 
HAWQ \cite{dong2019hawq} determines the bit-width for each block heuristically based on the top eigenvalue of Hessian. Unfortunately, both of them do not learn optimal bit-widths for heterogeneity. %are not based on learning optimal bit-widths for heterogeneity. 
% As a way of directly learning heterogeneity, 
Toward this, \cite{wang2019haq} and \cite{uhlich2020mixed} proposed a layer-wise heterogeneous quantization by exploiting reinforcement learning and learning dynamic range of quantizers, respectively. However, their results exhibit that almost all layers have up to 10-bit (at least 4-bit), which would be suboptimal. \cite{lou2020autoQ} presented a channel-wise heterogeneous quantization by exploiting hierarchical reinforcement learning, but channel-wise precision limits the structure of accelerators, thereby restricting the applicability of the model.

% \vspace{-0.1cm}
\section{Cluster-Promoting Quantization}\label{sec:method}

In this section, we first summarize the notations used in this paper and then present an overview of our method.

The variable $x$ denotes a weight or an activation of a full-precision network and $\widehat{x}$ indicates the quantized value of $x$. 
Here, we consider the following quantization grids for $x$: For a weight $x$, $\widehat{\mathcal{G}} \coloneqq [g_0, \ldots, g_{2^b-1}] = \alpha[-2^{b-1}, \ldots, 0, \ldots, 2^{b-1} - 1]$ where $b$ is the given bit-width and $\alpha > 0$ is a learnable parameter that controls the interval of quantization grids. For an activation $x$, quantization grids in $\widehat{\mathcal{G}}$ start from zero since the output of ReLU is always non-negative.  Lastly, $[n]$ denotes the set $\{0, 1, \cdots, n-1\}$ for a positive integer $n$.

%We first present a blueprint for our method before diving into the details.
Our main goal is to design a quantization algorithm that both finds the optimal $\alpha$ and clusters the underlying full-precision weights $x$ around quantization grids $\widehat{\mathcal{G}}$ cohesively in low bit-width regimes. % to preserve the performance of a full-precision network. %considering that the performance of a full-precision network can be preserved even after quantization if the underlying full-precision weights $x$ converge to the optimal quantization grids. 
As a neural network is overparametrized, there may exist a parameter such that the underlying full-precision weights $x$ crowd around some discrete values without performance degradation.
%while maintaining the performance of a full-precision network. 
Toward this, we propose the Cluster-Promoting Quantization (CPQ) that not only finds the optimal $\alpha$ but also helps the underlying full-precision weights $x$ congregate around quantization grids $\widehat{\mathcal{G}}$ cohesively. The proposed CPQ consists of two components: %which is the combination of a 
(i) a certain probabilistic parametrization for discretization (Section \ref{subsec:parametrization}) and (ii) our multi-class STE (Section \ref{subsec:multiclass_ste}). % is mainly characterized by Proposition \ref{prop} introduced in Section \ref{subsec:multiclass_ste}. 
Surprisingly, %we identify that without 
CPQ does not require any penalty or loss for clustering thanks to the combination of these two components as shown in Proposition \ref{prop} introduced in Section \ref{subsec:multiclass_ste}. %CPQ not only finds the optimal quantization grids but also helps the underlying full-precision weights $x$ congregate around quantization grids cohesively by Proposition \ref{prop} introduced in Section \ref{subsec:multiclass_ste}.}

%\paragraph{Notation.} \textcolor{blue}{Throughout this paper, $b$ is the bit-width given and $\alpha > 0$ is a learnable parameter that controls the interval between two adjacent quantization grids. $x$ denotes a network weight or an activation and $\widehat{x}$ indicates the quantized value of $x$. The variable $\epsilon$ represents the input noise for variational optmization \cite{Staines2012vo} that can follow any distribution with standard deviation $\sigma$.} % the possible noise in $x$ % Unless otherwise specified, the input noise $\epsilon$ can follow any distribution such as the uniform distribution or the normal distribution with zero mean and the standard deviation $\sigma$ that can be trained if necessary.}   

\subsection{Probabilistic Parametrization for Quantization}\label{subsec:parametrization}

% In order to allow gradient-based quantization, we introduce a parametrization in \cite{louizos2018relaxed} for our algorithmic purposes. 
% As a linear symmetric quantizer is hardware-friendly, we consider % Relaxed Quantization (RQ) considers the following quantization grids for weights: $\widehat{\mathcal{G}} = \alpha[-2^{b-1}, \ldots, 0, \ldots, 2^{b-1} - 1] \eqqcolon [g_0, \ldots, g_{2^b-1}]$. When quantizing activations, quantization grids in $\widehat{\mathcal{G}}$ start from zero since the output of ReLU is always non-negative. 
To permit gradient-based optimization, we assume that $x$ is perturbed by noise $\epsilon$ as $\widetilde{x} = x + \epsilon$. %, where $\epsilon$ can follow any distribution such as the uniform distribution or the normal distribution.
The variable $\epsilon$ represents random noise for variational optimization \cite{Staines2012vo} that can follow any distribution with zero mean and standard deviation $\sigma$. Here, let $\epsilon$ follow the logistic distribution $p(\epsilon) = \mathrm{Logistic}(0, \sigma)$ so that $p(\widetilde{x})$ is governed by  $\mathrm{Logistic}(x, \sigma)$. % where $\sigma$ represents the standard deviation of the logistic distribution. 
Under such $p(\widetilde{x})$, the \emph{unnormalized} probability of $\widetilde{x}$ being quantized to each quantization grid $g_i$ can be easily computed in a closed form as below:
\begin{align}\label{eqn:unnorm_prob}
    \pi_i % &= p(\widehat x = g_i | x, \alpha, \sigma) \\ %&= \mathbb{P}\big(\widetilde{x} \le (g_i + \alpha/2)\big) - \mathbb{P}\big(\widetilde{x} < (g_i - \alpha/2)\big)
    = \mathrm{Sigmoid}\Big({g_i + \frac{\alpha}{2} - x \over \sigma}\Big) - \mathrm{Sigmoid}\Big({g_i - \frac{\alpha}{2} - x \over \sigma}\Big), 
\end{align}
where the cumulative distribution function of the logistic distribution is a sigmoid function. Note that under \eqref{eqn:unnorm_prob}, $x$, $\alpha$, and $\sigma$ are trainable parameters. Given unnormalized categorical probabilities $\bm \pi = \{\pi_i\}_{i=0}^{2^b-1}$ for quantization grids $\widehat{\mathcal{G}}=\{g_i\}_{i=0}^{2^b-1}$, depending on how to design where $x$ is quantized according to $\bm \pi$, an algorithmic detail is determined. For instance, \cite{louizos2018relaxed} employed the Gumbel-Softmax trick \cite{Jang2016gumbel,Maddison2016concrete} based on $\bm \pi$. In this paper, we adopt such a probabilistic parametrization \eqref{eqn:unnorm_prob} from \cite{louizos2018relaxed} for our method.

%\subsection{Semi-Relaxed Quantization}\label{subsec:main} %  - Fixing Pitfalls of RQ

\subsection{Multi-Class Straight-Through Estimator}\label{subsec:multiclass_ste}

Given a probabilistic model for quantization like \eqref{eqn:unnorm_prob}, we define a new straight-through estimator (STE) as follows: % Let $r_i = {\pi_i} / {\sum_{j=0}^{2^b-1} \pi_j}$ where a way to compute $\pi_i$ can be changed depending on the distribution of $\epsilon$. Then
% \textcolor{blue}{In network quantization, it is crucial to cluster network parameters around quantization grids while maintaining the performance of full-precision network. However, adding an extra regularization or loss for clustering may incur performance degradation rather. Without any regularization or loss for clustering, on the basis of $\bm \pi$ parametrized by \eqref{eqn:unnorm_prob}, we propose the following multi-class straight-through estimator (STE) that theoretically and empirically demonstrates the superiority of clustering.} % As we discuss in Section \ref{subsec:parametrization}, how to design a variable $\bm r$ really matters, so we instead propose a following multi-class straight-through estimators (STE) that theoretically guarantees parameter clustering while avoiding unnecessary regularization:
\begin{align}
 	\textbf{Forward: }& y = \mathtt{one\_hot}[\argmax_{i} \pi_i] \label{eq:one_hot} \\ % & r_i = \frac{\pi_i}{\sum_{j=0}^{2^b-1} \pi_j} \text{ for } i \in [2^b-1] \label{eq:normalization} \\
 	\textbf{Backward: }& \frac{\partial \mathcal{L}}{\partial \pi_{i_{\text{max}}}} = \frac{\partial \mathcal{L}}{\partial y_{i_{\text{max}}}} ~\text{ and }~ \frac{\partial \mathcal{L}}{\partial \pi_i} = 0 \text{ for } \forall \, i \neq i_{\text{max}}, \label{eq:ste} 
\end{align}
where $y_i$ is the $i$-th entry of the one-hot vector $y$ and $\mathcal{L}$ is the cross entropy between the true label and the prediction made by a quantized neural network by the forward pass \eqref{eq:one_hot}. We dub \eqref{eq:one_hot} and  \eqref{eq:ste} the `\emph{multi-class STE}'.
%
% Now, we provide the concrete description of our multi-class STE. 
That is, in the forward pass, we directly select the mode (or the most likely grid) of the categorical distribution parametrized by the probabilistic model. 
% in Figure \ref{fig:distributions}-(g).
%To be concrete, the unnormalized probability of $x$ being quantized to each grid can be viewed as ${\pi_i}$ for $i \in \{0, \cdots, 2^b-1\}$, where $\pi_i$ can be computed as in \eqref{eqn:unnorm_prob}. In this manner, selecting a quantization grid for $x$ can be thought of as sampling from the categorical distribution with categories $\widehat{\mathcal{G}}=\{g_i\}_{i=0}^{2^b-1}$ and the corresponding unnormalized probabilities $\bm \pi = \{\pi_i\}_{i=0}^{2^b-1}$. % as illustrated in Figure \ref{fig:distributions}-(g). 
%Then, the grid $g_{i_{\mathrm{max}}}$ with $i_{\mathrm{max}} = \argmax_i \pi_i$ would be the most reasonable speculation due to the highest unnormalized probability. We therefore choose $g_{i_{\mathrm{max}}}$ (i.e., the mode of the categorical distribution parametrized by $\bm \pi$) as $\widehat{x}$.} % in the forward pass.} %, entirely discriminated from Gumbel-Softmax which selects the argmax among samples from the concrete distribution. As a result, SRQ does not suffer from samples with large quantization error at all.
%
% The last essential part for SRQ 
On the other hand, in the backward pass, it is required to handle the non-differentiable $\argmax$ operator in computing $i_{\mathrm{max}}$. % To encourage $x$ to congregate around quantization grids $\widehat{\mathcal{G}}$ cohesively as well as 
To allow gradient-based optimization, we enable backpropagation through a non-differentiable \emph{categorical} sample 
by backpropagating only through the path corresponding to $i_{\mathrm{max}}$. 
%approximating ${\partial \mathcal{L}}/{\partial \pi_{i_{\mathrm{max}}}}$ to ${\partial \mathcal{L}}/{\partial y_{i_{\mathrm{max}}}}$ and ${\partial \mathcal{L}}/{\partial \pi_i}$ to zero for all $i \ne i_{\mathrm{max}}$. %, where $\mathcal{L}$ is the cross entropy between the true label and the prediction made by a quantized neural network % weights (and activations) 
% as described in the previous paragraph and $y_i$ is the $i$-th entry of the one-hot vector $y = \mathtt{one\_hot}[\argmax_i r_i]$. %The forward and backward passes of SRQ are summarized as follows.

Note that the proposed multi-class STE can be thought of as a natural extension of binary case \cite{Bengio2013ste}, but this work is the first in-depth study on multi-class STE in terms of network quantization. Although \cite{Jang2016gumbel} proposed slightly different heuristic estimator, ST GS, that uses the Gumbel-softmax trick with another straight-through estimator to bypass the non-differentiability of discrete random variables, ST GS does not have any justification in the context of network quantization.
On the other hand, our multi-class STE theoretically and empirically demonstrates the superiority of clustering by the following proposition, when $\pi_i$ is computed as in \eqref{eqn:unnorm_prob}, even without any regularization or loss for clustering. % $r_i = {\pi_i} / {\sum_{j=0}^{2^b-1} \pi_j}$ in \eqref{eq:normalization} % Before introducing detailed procedures, we first emphasize the important property of our multi-class STE characterized by the following proposition, which encourages network parameters to gather around quantization grids cohesively. % $r_i = {\pi_i} / {\sum_{j=0}^{2^b-1} \pi_j}$ from \eqref{eqn:unnorm_prob}, 

\begin{prop}\label{prop}
    Let $\mathcal{L}$ be a loss function calculated from a quantized neural network using \eqref{eqn:unnorm_prob} and \eqref{eq:one_hot}. Under the assumption that $\lvert{\partial \mathcal{L} \over \partial y_{i_{\text{max}}}}\rvert$ is bounded, the gradient of $\mathcal{L}$ with respect to full precision variable $x$ from \eqref{eq:ste}, ${\partial \mathcal{L} \over \partial x}$, converges to zero as a weight $x$ approaches its nearest quantization grid $g_{i_{\text{max}}}$. 
\end{prop}
%
% and $r_i = \pi_i$ from \eqref{eqn:unnorm_prob} for each $i$\footnotemark[1]
% \addtocounter{footnote}{1}
% \footnotetext[1]{As $i_{\text{max}}$ does not change under the assumption, it is not unreasonable.}
%
\begin{figure}[t]
	\begin{center}
	    \includegraphics[width=0.9\linewidth]{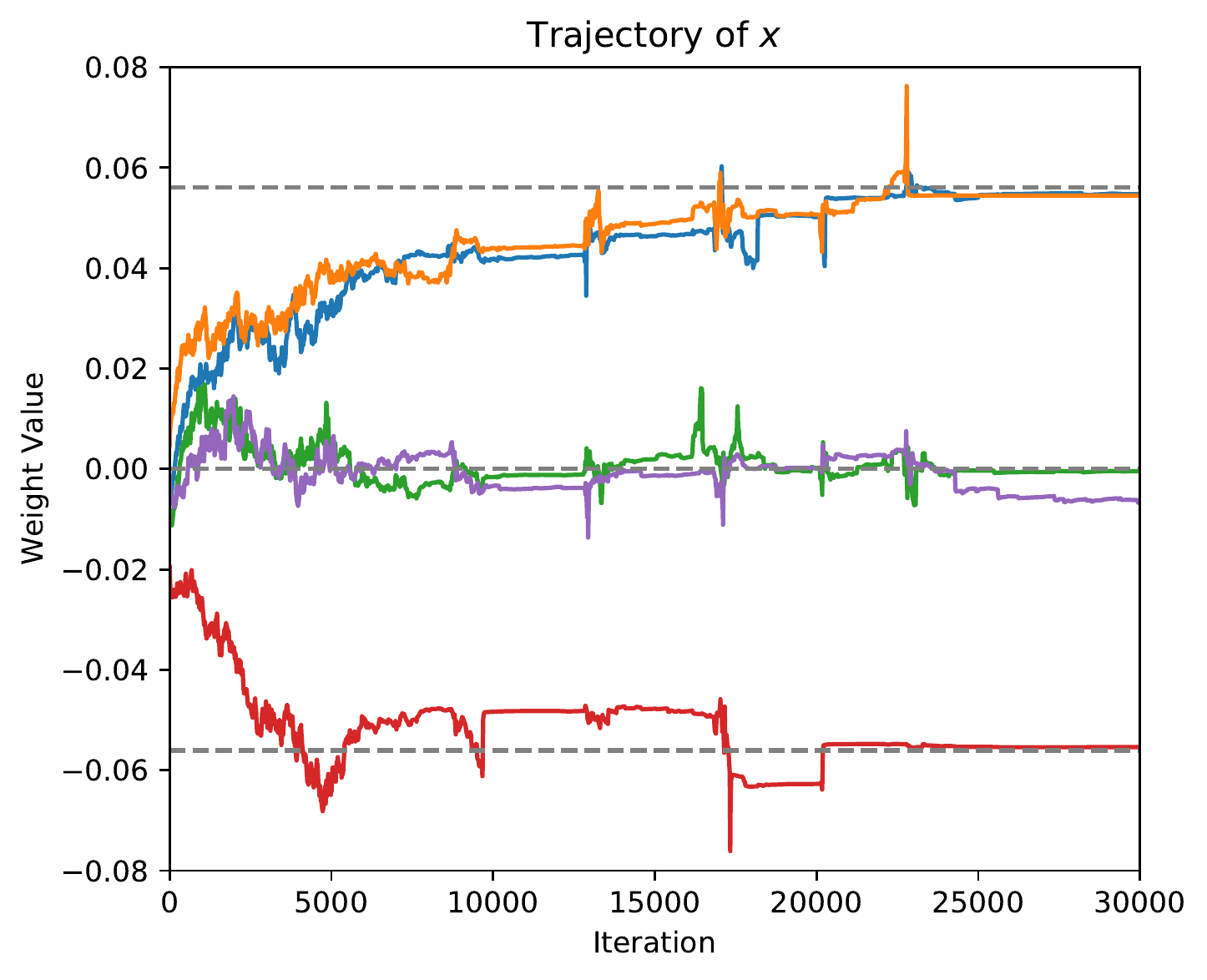}
	\end{center}
	\caption{Trajectories of five random weights in the second layer when training LeNet-5 on MNIST in $3$-bit. The $x$-axis indicates the number of training iterations, and the $y$-axis represents the value of weight. The horizontal dashed lines (gray) denote quantization grids after training.}
	\label{fig:weight_trajectory}
\end{figure}
\begin{figure}[t]
	\begin{center}
	    \includegraphics[width=0.9\linewidth]{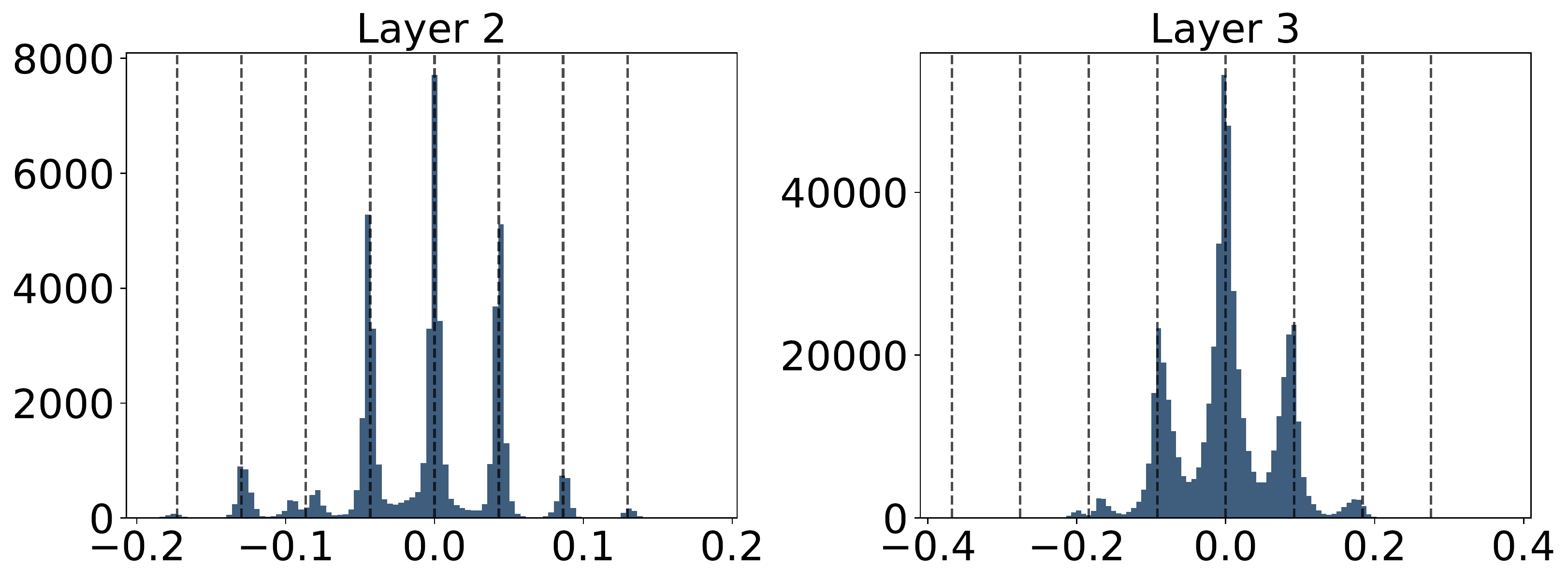}
	\end{center}
	\caption{Weight distributions for $3$-bit quantized LeNet-5 by our method, CPQ. The $x$-axis and $y$-axis indicate weight values and their frequencies, respectively. The vertical dashed lines denote quantization grids.} 
	\label{fig:weight_distribution}
	\vskip -3pt
\end{figure}
\noindent By Proposition \ref{prop}, once $x$ is trained to become near $g_{i_{\text{max}}}$, $x$ can be kept to stay around $g_{i_{\text{max}}}$ as seen in Figure \ref{fig:weight_trajectory}, thus making it possible to cluster the underlying full-precision weights around quantization grids cohesively as seen in Figure \ref{fig:weight_distribution}. As our multi-class STE with \eqref{eqn:unnorm_prob} can be qualitatively distinct from other unjustified estimators from this perspective, we call the combination of \eqref{eqn:unnorm_prob} and the multi-class STE `Cluster-Promoting Quantization (CPQ)'. The overall procedure of CPQ is described in Algorithm \ref{alg:cpq_training}.

\begin{figure*}[t]
    \vskip -6pt
	\centering
	\subfigure{\includegraphics[width=0.7\linewidth]{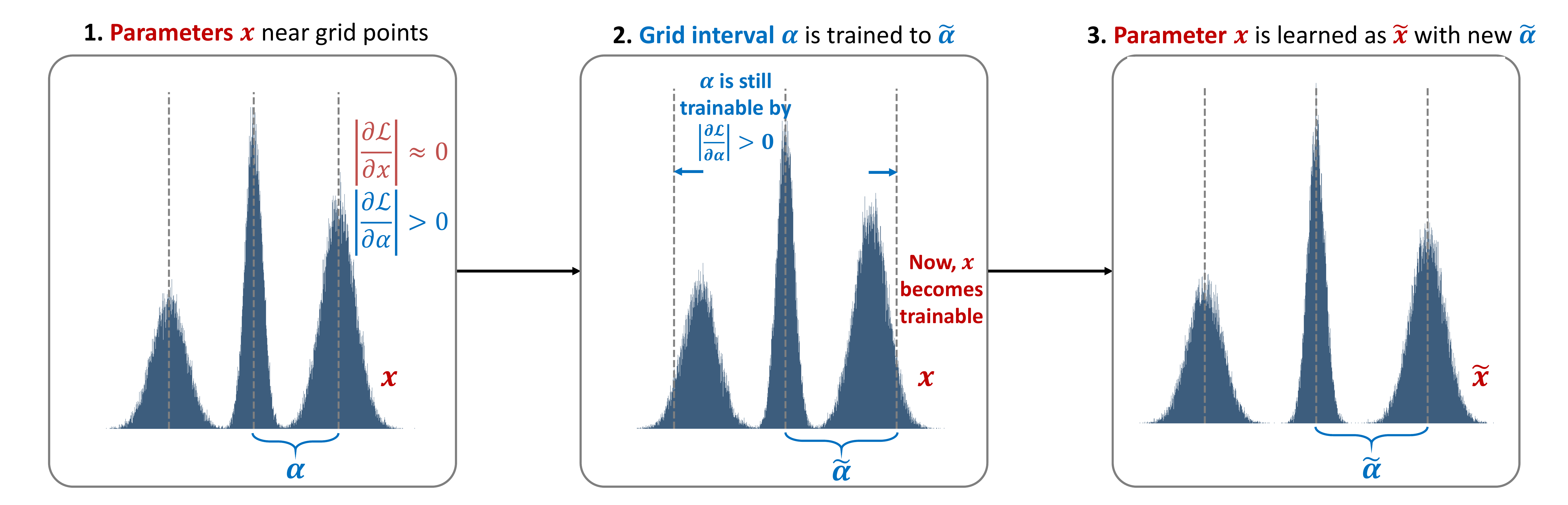}}
	\caption{Training procedure when weights are close to quantization grids.}
	\label{fig:training_procedure}
% 	\vskip -8pt
\end{figure*}

One might wonder that the almost zero gradient near quantization grids may make a network untrainable, which would not be a gradient-based learning. Although ${\partial \mathcal{L} \over \partial x}$ is almost zero when $x$ is close to $g_{i_{\text{max}}}$, $\alpha$ is still trained to find the better grid points. After $\alpha$ is updated, if the gap between $x$ and $\alpha$ is widened , then $x$ is trained accordingly. Hence, a network will continue to train until it finds the optimal $\alpha$. Such a training procedure is illustrated in Figure \ref{fig:training_procedure}. % Notice that $x$ is close to $0.0$ at first, but it is consequently clustered around $\alpha$. This seems primarily due to the fact that $\lvert{\partial \mathcal{L} \over \partial y_{i_{\text{max}}}}\rvert$ can be so large in the beginning of training that the updated weight may pass by the neighborhood of $0.0$.

In addition to Proposition \ref{prop}, our multi-class STE has another strength: it makes the variance of gradients become indeed zero, which has to do with what \cite{louizos2018relaxed} highlighted to train a network with low bit-widths successfully. As our multi-class STE always chooses the mode of the categorical distribution paramterized by a probabilistic model (i.e., there is no randomness in the forward pass \eqref{eq:one_hot}) and the gradient of $\mathcal{L}$ with respect to the individual categorical probabilities is exactly \emph{zero} everywhere except for the coordinate corresponding to the mode in the backward pass \eqref{eq:ste}, the variance of our gradient estimator becomes zero.

\begin{algorithm}[t]
\small
 	\caption{\textbf{C}luster-\textbf{P}romoting \textbf{Q}uantization (CPQ)}
 	\label{alg:cpq_training}
 	\begin{algorithmic}[1]
 		\State {\bfseries Input:} Training data $\mathcal{D}$, network parameters $\{W_l, b_l\}_{l=1}^{L}$, layer-wise grid interval parameters and the standard deviations of a logistic distribution in the $l$-th layer $\{\alpha_l, \sigma_l\}_{l=1}^{L}$.
 		\State {\bfseries Output:} A low bit-width model with quantized network parameters $\{\widehat W_l, \widehat b_l\}_{l=1}^{L}$ after deployment procedure.
 		\State {\bfseries Initialize:} Bit-width $b$ and parameters $\{W_l, b_l, \alpha_l, \sigma_l\}_{l=1}^{L}$. Initialize layer-wise grid $\widehat{\mathcal{G}}_l \coloneqq [g_{l,0}, g_{l,1}, \cdots, g_{l, 2^b-1}] = \alpha_l [-2^{b-1}, \cdots, 2^{b-1} - 1]$ for $l \in \{1, \cdots, L\}$. 
        \Procedure{Training}{}
        \State // ~Forward pass
        \For{$l = 1, \cdots, L$}
            \Let {$x$}{Each entry of $W_l$ or $b_l$}
            \State {$I_l = \widehat{\mathcal{G}}_l - \alpha/2$}
            \Comment{Shift the grid by $-\alpha/2$}
    		\State {$F = \text{Sigmoid}\Big(\frac{I_l - x}{\sigma_l}\Big)$} \Comment{Compute CDFs}
    		\State {$\pi_i = {F[i+1] - F[i]} \,\, \text{for } i \in [2^b-1]$} \Comment{Eq. \eqref{eqn:unnorm_prob}}
    		\State $y = \mathtt{one\_hot}[\argmax_i \pi_i]$ \Comment{Eq. \eqref{eq:one_hot}}
    		\State $\widehat x = y \odot \widehat{\mathcal{G}_l}$ \Comment{Quantization}
    		%\Let{$\widehat{W}_l$}{Each entry of $W_l$ quantized to $\widehat x$}
    		%\Let{$\widehat{b}_l$}{Each entry of $b_l$ quantized to $\widehat x$}
    		%\vspace{+0.2cm}
    		%\State Forward pass with quantized $\widehat W_l$ and $\widehat b_l$
    		\State Activation can be quantized in the same way
        \EndFor
        \State // ~Backward pass
        \For{$l = L, \cdots, 1$}
            \State Compute gradients $(\frac{\partial \mathcal{L}}{\partial W_l}, \frac{\partial \mathcal{L}}{\partial b_l}, \frac{\partial \mathcal{L}}{\partial \alpha_l}, \frac{\partial \mathcal{L}}{\partial \sigma_l})$ \Comment{Eq. \eqref{eq:ste}}
            \State Update parameters $(W_l, b_l, \alpha_l, \sigma_l)$
        \EndFor
		\EndProcedure
		\Procedure{Deployment}{}
		    \For{$l = 1, \cdots, L$}
		        \State $\widehat W_l = \min(\max(\alpha_l \cdot \mathrm{Round}(W_l/\alpha_l), g_{l,0}), g_{l,2^b-1})$
		        \State $\widehat b_l = \min(\max(\alpha_l \cdot \mathrm{Round}(b_l/\alpha_l), g_{l,0}), g_{l,2^b-1})$
		    \EndFor
		\EndProcedure
 	\end{algorithmic}
\end{algorithm}

% \vspace{-0.1cm}
\section{DropBits and Its Extension to Heterogeneous Quantization}
We propose a novel bit-drop technique named \emph{DropBits} to reduce the bias of the multi-class STE (Section \ref{subsec:dropbits}). We also impose an extra regularization on DropBits to permit heterogeneous quantization (Section \ref{subsec:learning_bit}) and put forward a new hypothesis for quantization (Section \ref{subsec:lottery}).

\subsection{DropBits}\label{subsec:dropbits}

Although our multi-class STE enjoys zero variance of gradients, %extremely small,
it is biased to the mode as the binary one in \cite{Bengio2013ste}. To reduce the bias of STE, \cite{chung2016hierarchical} propose the slope annealing trick, but this strategy is only applicable to the binary case. To address this limitation, we propose a novel bit-drop method, \emph{DropBits}, to decrease the bias of our multi-class STE. Inspired by dropping neurons in Dropout \cite{srivastava2014dropout}, we \emph{drop an arbitrary number of grid points at random every iteration}, where in effect the probability of being quantized to dropped grid points becomes zero. 

\begin{figure}[t]
	\vskip -5pt
	\centering
	\subfigure[Endpoints-sharing mask]{\includegraphics[width=0.45\linewidth]{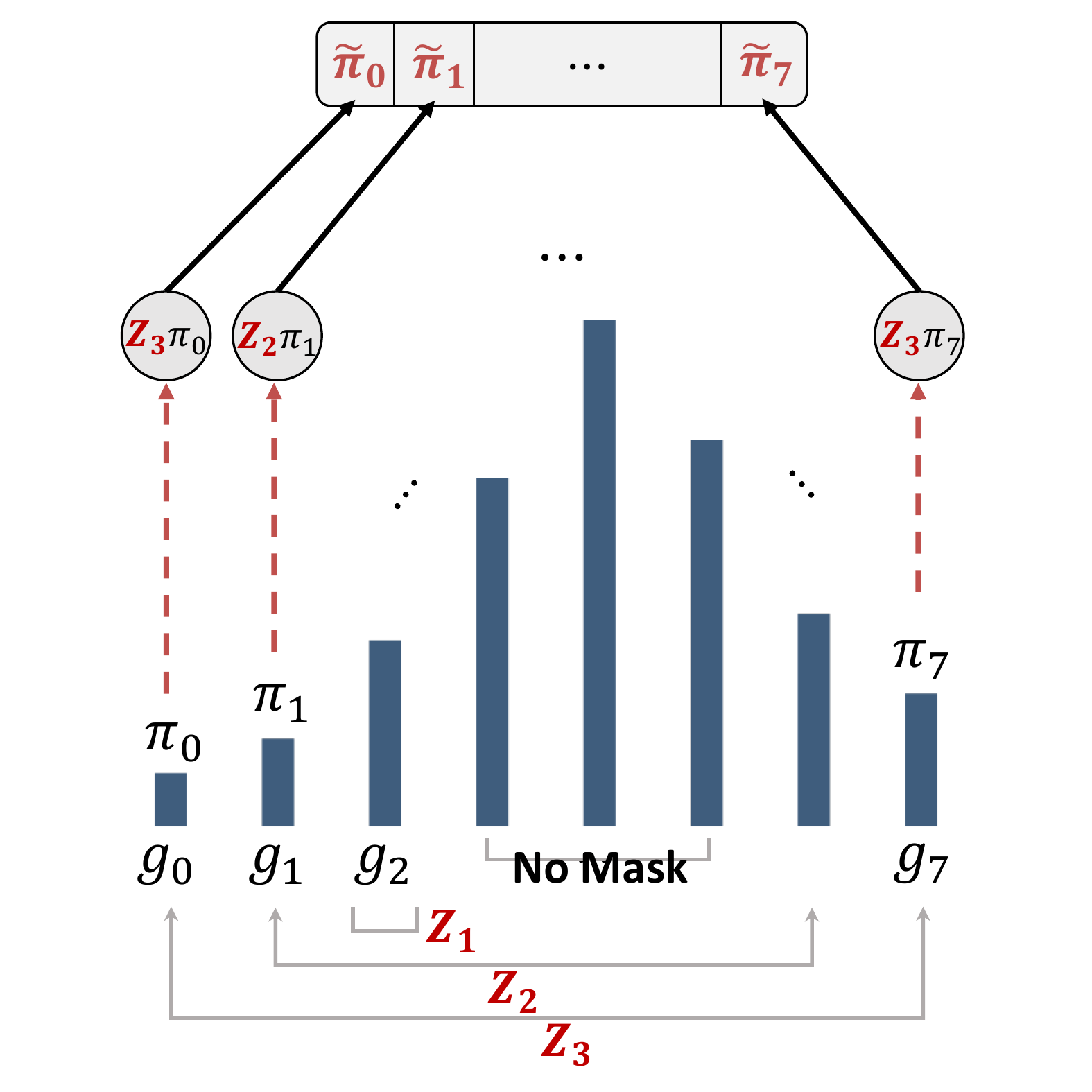}}
	\subfigure[Bitwise-sharing mask]{\includegraphics[width=0.45\linewidth]{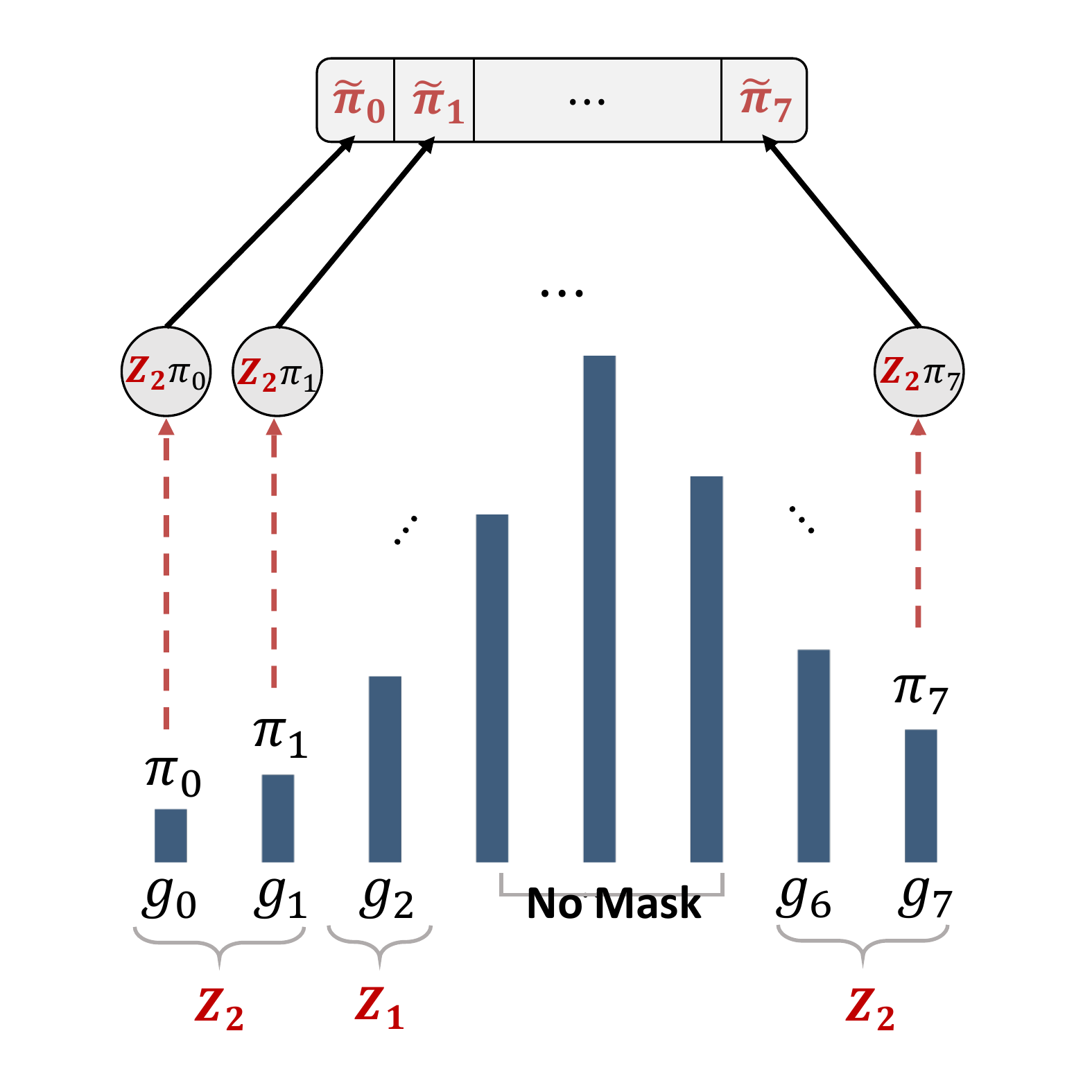}}
	\caption{Designs of two masks for 3-bit}
	\vskip -5pt 
	\label{fig:mask_design}
\end{figure}

However, the design policy that each grid point has its own binary mask would make the number of masks increase exponentially with bit-width. Taking into consideration appropriate noise levels with a less aggressive design, the following two examples are available: \textbf{(a)} endpoints in the grids share the same binary mask, and \textbf{(b)} the grid points in the same bit-level share the same binary mask (see Figure \ref{fig:mask_design}). Hereafter, we consider \textbf{(b)} bitwise-sharing masks for groups of grid points, unless otherwise specified.

Now, we introduce how to formulate binary masks. Unlike Dropout implementation through dividing activations by $1 - p$ (here, $p$ is a dropout probability), we employ an explicit binary mask $Z$ whose probability $\Pi$ can be optimized jointly with model parameters. The Bernoulli random variable being non-differentiable, we relax a binary mask via the \emph{hard concrete} distribution \cite{louizos2017l0reg}. While the binary concrete distribution \cite{Maddison2016concrete} has its support $(0, 1)$, the hard concrete distribution stretches it slightly at both ends, thus concentrating more mass on exact $0$ and $1$. Assuming disjoint masks, we describe the construction of a binary mask $Z_k$ for the $k$-th bit-level using the hard concrete distribution as follows.
% \vspace{-0.1cm}
\begin{align} \label{eq:mask}
	&U_k \sim \text{Uniform}(0, 1), \\
	&S_k = \text{Sigmoid}\Big(\big(\log{U}_k\!-\!\log{(1 - U_k)} + \log {\Pi_k \over 1-\Pi_k}\big) / \tau' \Big) \nonumber \\
	&\bar{S}_k  = S_k(\zeta - \gamma) + \gamma  \quad \text{and}\quad
	Z_k  = \min(\max(\bar{S}_k, 0), 1) \nonumber
	 % \,\, \text{for} \,\, k = 0, \cdots, 2^b-1 
	%\mathcal{L}_{\text{reg}} & = {\sum_{i=0}^{b-2} \ceil[\big]{Z_{b-2-i}}}\left(\prod_{j=0}^{i-1}\left(1-\ceil[\big]{Z_{b-2-j}}\right)\right)(1-Q_{\bar{s}_i}(0))
\end{align}
% \vspace{-0.2cm}
where $\tau'$ is a temperature for the hard concrete distributions with $\gamma < 0$ and $\zeta > 0$ reflecting stretching level. For $i = 2^{b-1}-1, 2^{b-1}$ and $2^{b-1}+1$, we do not sample from the above procedure but fix $Z = 1$ to prohibit all the binary masks from becoming zero (see `No Mask' in Figure \ref{fig:mask_design}). % \textcolor{blue}{k is bit index} 
%, that is, $\pi_k$ becomes the probabilities of being quantized to $-\alpha, 0$ and $\alpha$ 
% $m$ is the index of a binary mask corresponding to $\pi_k$ as shown in Figure \ref{fig:mask_design}, 

%Under this construction, we now give detailed descriptions for two mask designs, \textbf{(i)} and \textbf{(ii)}. The Figure \textcolor{blue}{The Figure \ref{fig:mask_design} shows the detailed DropBits procedure.} 

\begin{figure*}[t]
    \vskip -6pt
	\centering
	\subfigure[Categorical distribution]{\includegraphics[width=0.25\linewidth]{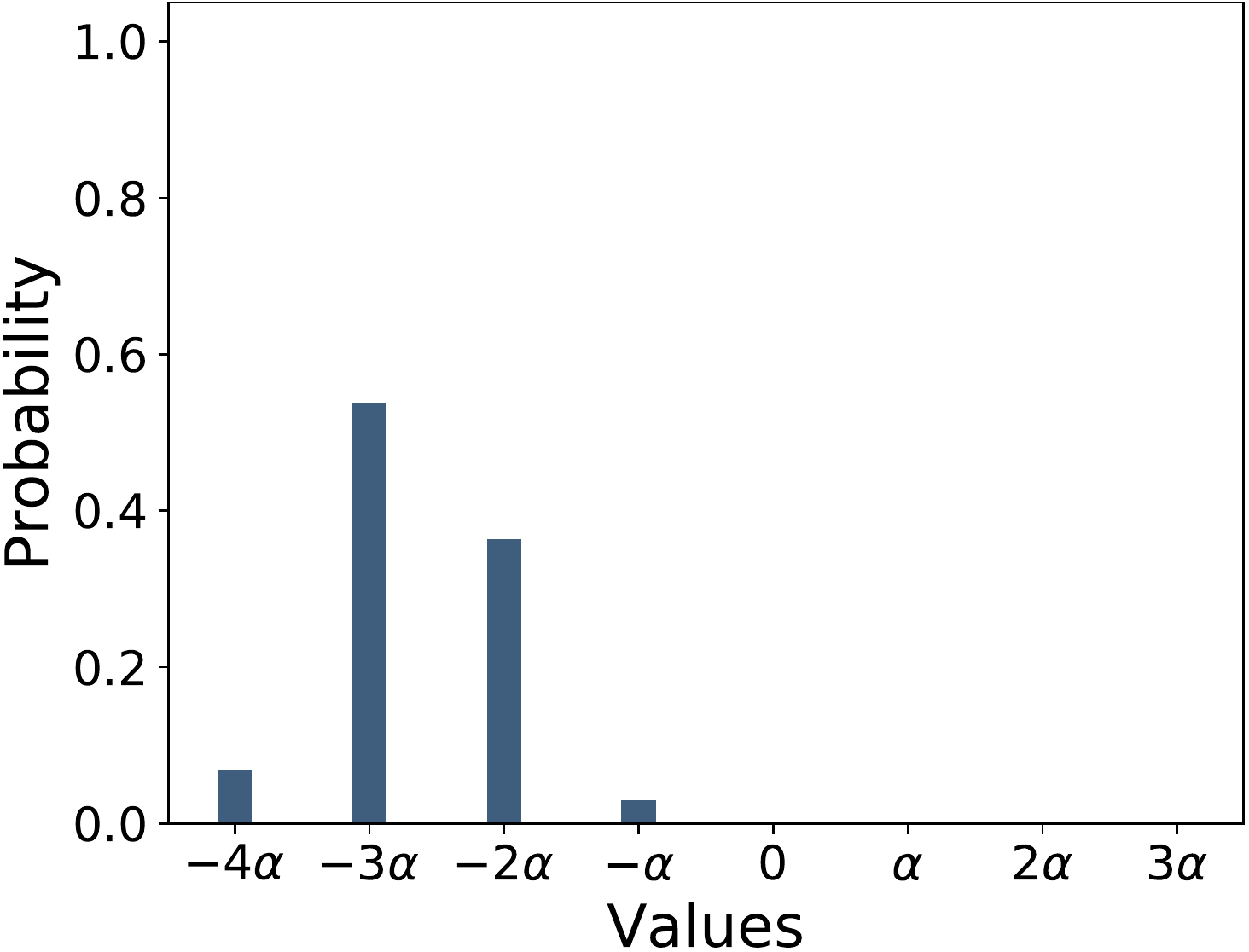}}
	\subfigure[CPQ]{\includegraphics[width=0.25\linewidth]{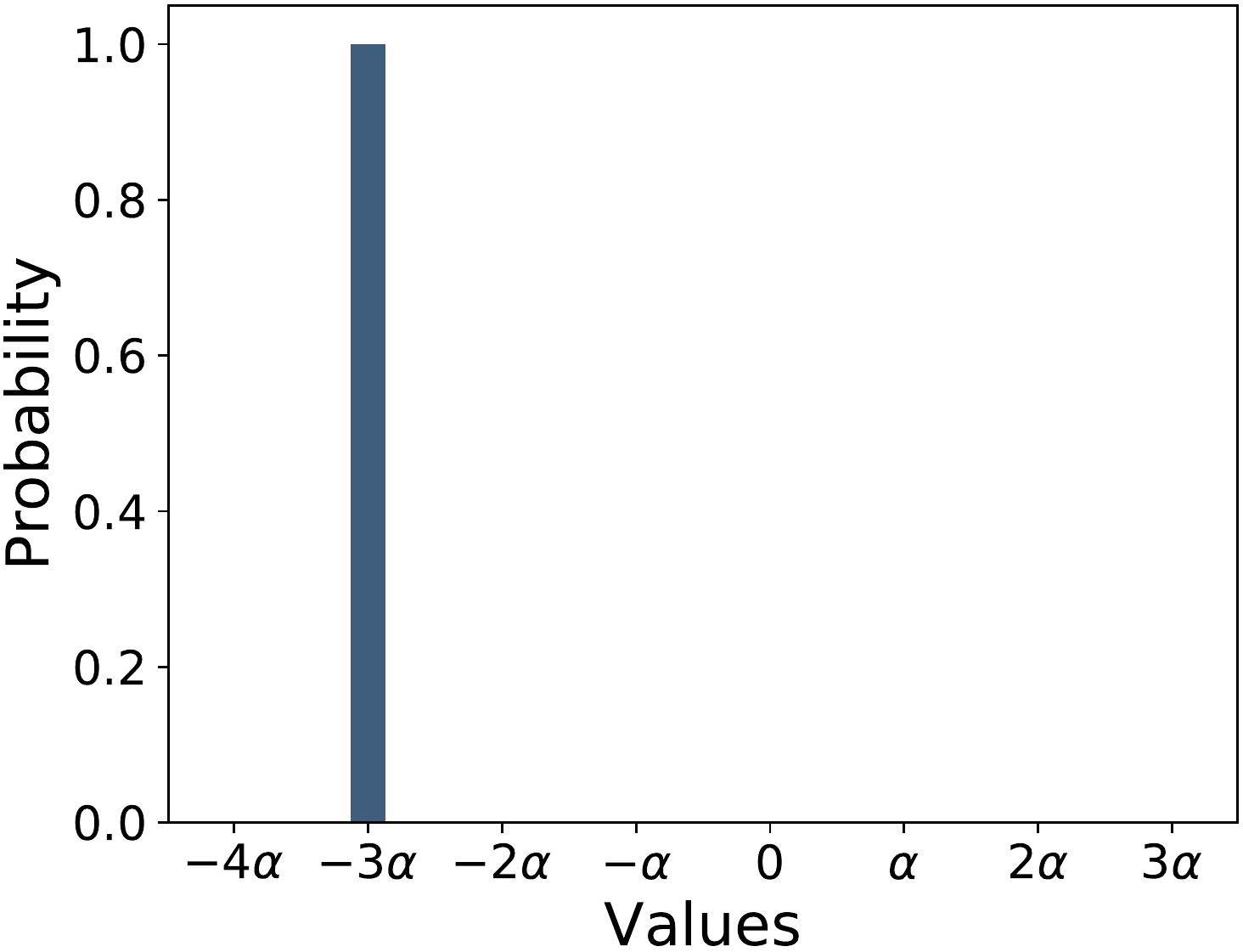}}
	\subfigure[CPQ + DropBits]{\includegraphics[width=0.25\linewidth]{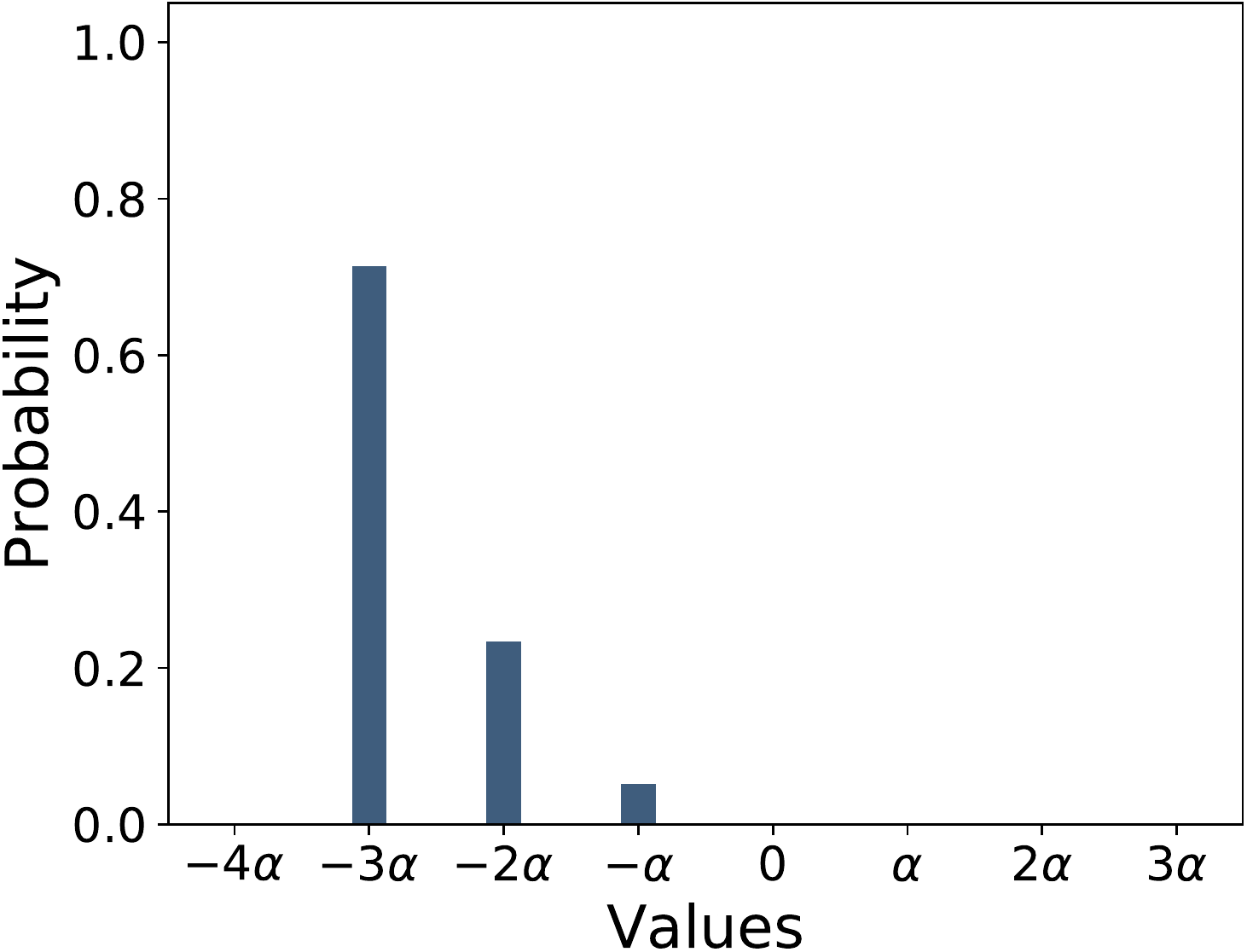}}
% 	\vskip -12pt
	\caption{The illustration of the effect of DropBits on CPQ. For a certain weight, \textbf{(a)} the categorical distribution indicates $\pi_i/\Sigma_{j=0}^{7}\pi_j$ for each grid $(i=0,\cdots,7)$, \textbf{(b)} the distribution of CPQ is a sampling distribution after taking the argmax of $\pi_i$, and \textbf{(c)} the distribution of CPQ + DropBits is a sampling distribution after taking the argmax of $\widetilde{\pi}_i$. Here, $\Pi_k$'s are initialized to $0.7$ for clear understanding.}
	\label{fig:compare_dist}
\end{figure*}

\begin{figure*}[t]
	\centering
	\subfigure{\includegraphics[width=\linewidth]{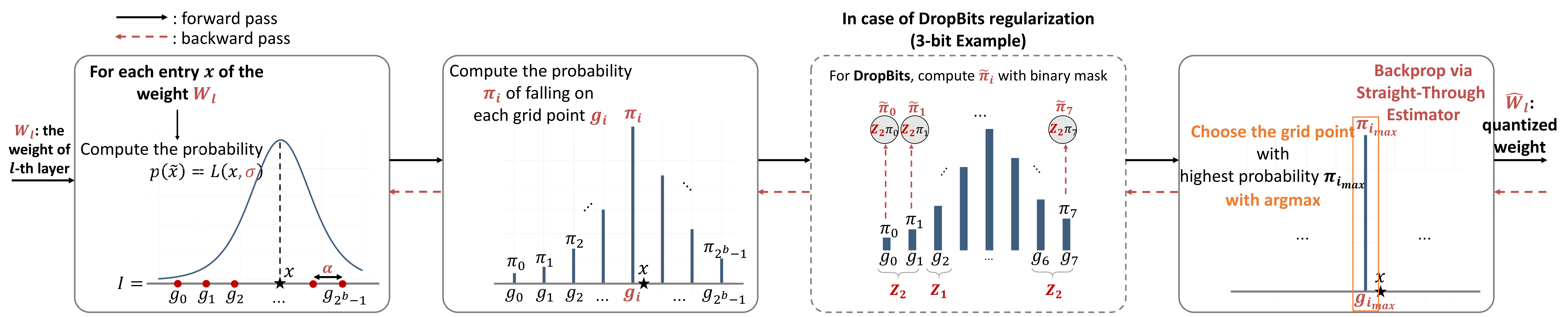}}
	\vskip 5pt
	\caption{Illustration of Cluster-Promoting Quantization (CPQ) framework with DropBits technique.}
	\label{fig:srq}
	\vskip -3pt
\end{figure*}

With the value of each mask generated from the above procedure, the probability of being quantized to any grid point is re-calculated by multiplying $\pi_i$'s by their corresponding binary masks $Z_k$'s (e.g. $\widetilde{\pi}_0 = Z_2 \cdot \pi_0$ in Figure \ref{fig:mask_design} (b))
% (that is, $\widetilde{\pi} = Z \odot \pi$) 
and then normalizing them (to sum to 1).
% by adjusting $\pi_i$'s to $\widetilde{\pi}_i$'s based on $Z_k$'s
As seen in Figure \ref{fig:compare_dist}, the sampling distribution of CPQ is biased to the mode, $-3\alpha$. For an appropriate value of $\Pi_k$, the sampling distribution of CPQ + DropBits can more resemble the original categorical distribution than that of CPQ by adjusting $\pi_i$'s to $\widetilde{\pi}_i$'s based on $Z_k$'s via DropBits, which means that DropBits is able to reduce the bias of the multi-class straight-through estimator in CPQ effectively. %, the distribution of samples with DropBits describes the original categorical distribution better %is more similar to the original categorical distribution 
% than that without DropBits, which means that DropBits could effectively reduce %is capable of reducing 
% the bias of the multi-class straight-through estimator in SRQ. 
Not only that, DropBits does not require any hand-crafted scheduling at all due to the learnable characteristic of $\Pi_k$, %since $\Pi_k$'s are learnable
%and $Z_k$'s go to one in most cases as training progresses, 
whereas such scheduling is vital for Gumbel-Softmax \cite{Jang2016gumbel, Maddison2016concrete} and slope annealing trick \cite{chung2016hierarchical}. %requires annealing scheduling 
%for successful training.} Note that DropBits is only applied during training not to make any burden in inference time.

Although quantization grids for weights are symmetric with respect to zero, those for activations start from zero, which makes it difficult to exploit symmetrically-designed DropBits for activations. Therefore, DropBits is applied only for weights in our experiments. Assuming that $Z_k$'s are shared across all weights of each layer, the overall procedure is described in Figure \ref{fig:srq}. The overall algorithm of CPQ + DropBits is deferred to Appendix due to space limit. %  \ref{alg:srq_training} 

\subsection{Learning Bit-width towards Resource-Efficiency}\label{subsec:learning_bit}

As noted in Section \ref{sec:intro} and \ref{sec:related}, recent studies on heterogeneous quantization use at least 4-bit, up to 10-bit, which leaves much room for saving %more consumption of 
energy and memory. %on the contrary to the goal of quantization. 
Towards more resource-efficient method, we introduce an additional regularization on DropBits to drop redundant bits.

As the mask design in Figure \ref{fig:mask_design}-(b) reflects the actual bit-level and the probability of each binary mask in DropBits is learnable, we can penalize the case where we use higher bit-levels via a sparsity encouraging regularizer like $\ell_1$. % such as $\ell_1$ \cite{tibshirani1996regression}. 
As \cite{louizos2017l0reg} proposed a relaxed $\ell_0$ regularization using the hard concrete binary mask, we adopt this continuous version of $\ell_0$ as a sparsity inducing regularizer. Following \eqref{eq:mask}, we define the smoothed $\ell_0$-norm as
%\begin{align}\label{eq:l0reg}
$\mathcal{R}(Z; \Pi) = \mathrm{Sigmoid}(\log \frac{\Pi}{1-\Pi} - \tau' \log \frac{-\gamma}{\zeta})$.
%\end{align}
One caveat here is that we do not have to regularize masks for low bit-level if a higher bit-level is still alive (in this case such a high bit-level is still necessary for quantization). % , regardless of the values of low bit-level masks
We thus design a regularization in such a specific way as only to permit the probability of a binary mask at the current highest bit-level to approach zero. More concretely, for bit-level binary masks $\{Z_k\}_{k=1}^{b-1}$ as in Figure \ref{fig:mask_design}-(b) and the corresponding probabilities $\{\Pi_k\}_{k=1}^{b-1}$, our regularization term to learn the bit-width is % can be formulated as % follows:  
% \vspace{-0.1cm}
\begin{align}
    &\mathcal{R}\big(\{Z_k\}_{k=1}^{b-1}, \{\Pi_k\}_{k=1}^{b-1}\big) \nonumber\\
    &= \sum\limits_{k=1}^{b-1} \mathbb{I}(Z_{k} > 0) \Big(\prod\limits_{j=k+1}^{b-1}\mathbb{I}(Z_j = 0)\Big)  \mathcal{R}(Z_k; \Pi_k).
\end{align}
% \vspace{-0.1cm}

Note that $\{Z_k\}_{k=1}^{b-1}$ is assigned to each group (e.g. all weights or activations in a layer or channel for instance). Hence, every weight in a group shares the same sparsity pattern (and bit-width as a result), and learned bit-widths across groups are allowed to be heterogeneous. 

% \vspace{-0.1cm}
Assuming the $l$-th \emph{layer} shares binary masks $\bm Z^l \coloneqq \{Z^l_k\}_{k=1}^{b-1}$ associated with probabilities $\bm \Pi^l \coloneqq \{\Pi^l_k\}_{k=1}^{b-1}$, our final objective function for a $L$-layer neural network becomes 
%\begin{align*}
	$\mathcal{L}(\bm \theta, \bm \alpha, \bm \sigma, \bm Z, \bm \Pi) + \lambda \sum_{l=1}^{L} \mathcal{R}(\bm{Z}^l, \bm{\Pi}^l)$,
%\end{align*}
where $\bm \alpha = \{\alpha_l\}_{l=1}^{L}$ and $\bm \sigma = \{\sigma_l\}_{l=1}^{L}$ represent the layer-wise grid interval parameters and standard deviations of logistic distributions, $\bm Z = \{\bm Z^l\}_{l=1}^L$, $\bm \Pi = \{\bm \Pi^l\}_{l=1}^L$, and $\lambda$ is a regularization parameter. In inference phase, we just drop unnecessary bits based on the values of $\bm \Pi$. %Here, we can achieve layer-wise \emph{heterogeneity} by controlling $\lambda$ value.

\begin{table}[t]
    \vskip -6pt
	\centering
	{
	\caption{Test error ($\%$) for LeNet-5 on MNIST and VGG-7 on CIFAR-10. ``Ann.'' stands for annealing the temperature of the Gumbel-Softmax trick in RQ.}
	\footnotesize
	\begin{tabular}{c|c|cc|cc}
	\toprule
	Dataset & \makecell{\# Bits \\ W./A.} & RQ & \makecell{RQ + \\ Ann.\footnotemark[2]} & CPQ & \makecell{CPQ + \\DropBits} \\
	\midrule
	& $4/4$ & $0.58$ & $0.62$ & $0.59$ & $\mathbf{0.53}$ \\
	MNIST & $3/3$ & $0.69$ & $0.74$ &  $0.67$ & $\mathbf{0.58}$ \\
	& $2/2$ & $0.76$ & $-$ & $0.72$ & $\mathbf{0.63}$ \\
	\midrule
	& $4/4$ & $8.43$ & $8.47$ & $7.15$ & $\mathbf{6.85}$ \\
	CIFAR-10 & $3/3$ & $9.56$ & $10.78$ & $7.08$ & $\mathbf{6.94}$ \\
	& $2/2$ & $11.75$ & $-$ & $7.68$ & $\mathbf{7.51}$ \\
	\bottomrule 
	\end{tabular}
	\label{tab:ablation}
	}
\end{table}
\addtocounter{footnote}{2}
\footnotetext[2]{We cannot reproduce the results of RQ in 2-bit, so we experiment only on 3-bit and 4-bit RQ}

\begin{figure}[t]
    % \vskip -12pt
	\centering
	\subfigure{\includegraphics[width=0.7\linewidth]{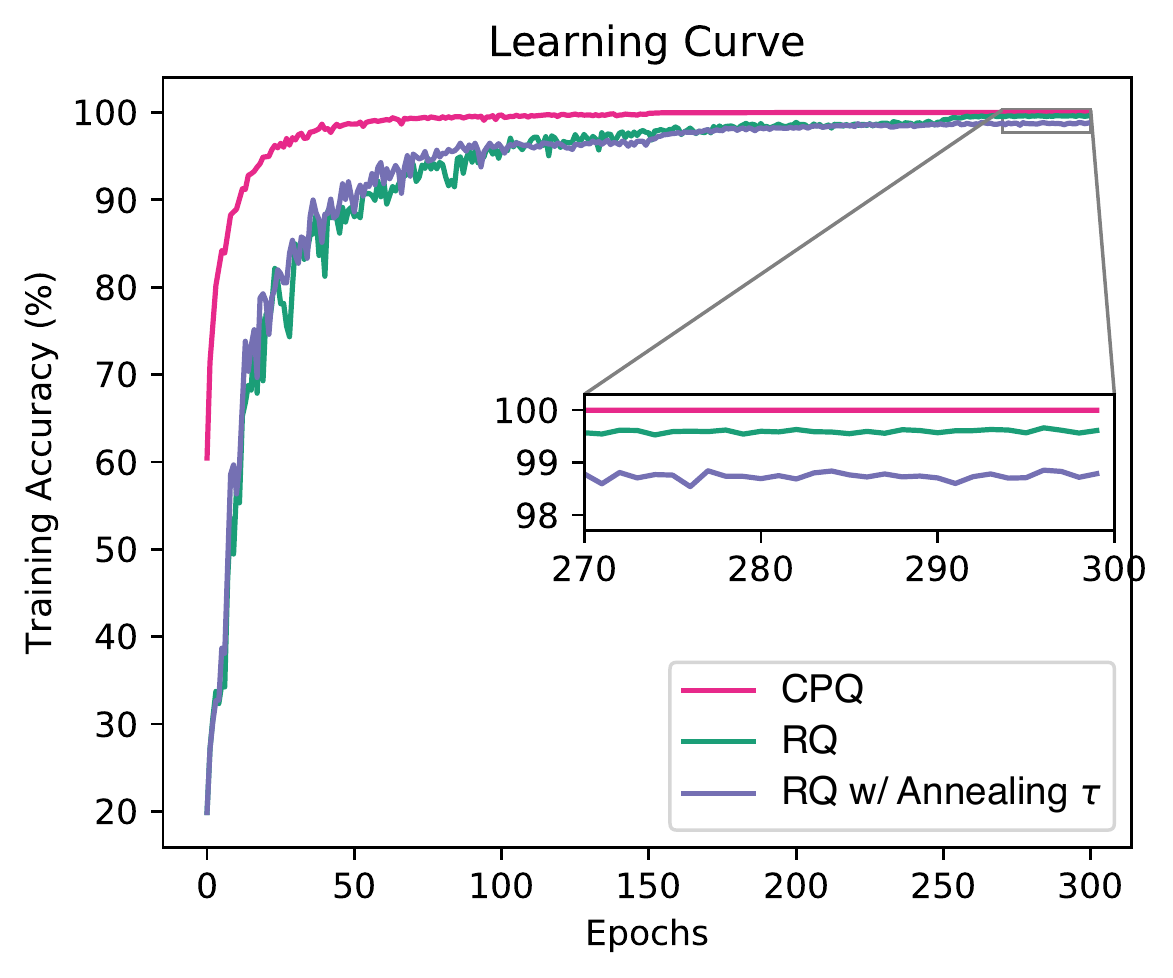}}
% 	\vskip -12pt
	\caption{Learning curves of VGG-7 quantized by RQ, RQ with annealing $\tau$, and CPQ in $3$-bit.}
	\label{fig:learning_curves}
	\vskip -5pt
\end{figure}

% \vspace{-0.2cm}
\subsection{New Hypothesis for Quantization}\label{subsec:lottery}

% \vspace{-0.2cm}
%Recently, \cite{frankle2018the} articulate the ``lottery ticket hypothesis'' which states that one can find sparse sub-networks, which they term ``winning tickets'', via network pruning. Motivated by this work, we similarly propose \emph{quantized lottery ticket hypothesis}.
\cite{frankle2018the} articulated the `lottery ticket hypothesis', stating that one can find some sparse sub-networks, `winning tickets', from randomly-initialized, dense neural networks that are easier to train than sparse networks resulting from pruning. %. To find the winning tickets, they employ a magnitude-based pruning which alternates between: (i) training the network, (ii) pruning fixed portion of the parameters with low magnitude, and (iii) resetting the remaining parameters back to their original initialization. The authors experimentally validate this conjecture on several benchmark architectures and datasets by identifying that such a sub-network really exists.
In this section, we define a new hypothesis for quantization with slightly different (opposite in some sense) perspective from the original one.
% \textcolor{blue}{similarly is enough?} 

% \vspace{-0.15cm}
% Toward this, we define the following notation: $\succ_{\mathrm{bit}}$ and $=_{\mathrm{bit}}$, where 
\textbf{Notation.} $a \succ_{\mathrm{bit}} b$ and $a =_{\mathrm{bit}} b$ denote that $a$ has strictly higher bit-width than $b$ for at least one of all groups (e.g. channels or layers), and $a$ has the same bit-precision as $b$ across all groups, respectively.

% \vspace{-0.15cm}
\textbf{Definition.}
For a network $f(x; \theta)$ with randomly-initialized parameters $\theta$, let $f(x; \theta')$ be a quantized network from $f(x; \theta)$ such that $\theta \succ_{\mathrm{bit}} \theta'$. If the accuracy of $f(x; \theta')$ is higher than that of $f(x; \theta'')$ where $f(x; \theta'')$ is trained from scratch with fixed bit-widths such that $\theta' =_{\mathrm{bit}} \theta''$, $f(x; \theta')$ is referred to as a $\emph{quantized sub-network}$ of $f(x; \theta)$.

%\textbf{Quantized Lottery Ticket Hypothesis.} 
%For a network $f(x; \theta)$ with randomly-initialized network parameters $\theta$ with $b$-bit precision, there exists a reduced network $f(x; \theta')$, which we term ``quantized winning ticket'', such that $\theta \succeq_{\mathrm{bit}} \theta'$ where $a \succeq_{\mathrm{bit}} b$ means that $a$ has a higher bit-precision than $b$. A quantized winning ticket outperforms the network trained from scratch with learned bit-structure.

% \vspace{-0.1cm}
This hypothesis implies learning bit-width would be superior to pre-defined bit-width. To the best of our knowledge, our study is the first attempt to delve into this hypothesis. % All previous work on heterogeneous quantization does not check this hypothesis, but it is critical to check this hypothesis because an algorithm would be useless if the algorithm cannot find a quantized winning ticket. We therefore apply this hypothesis to our heterogeneous quantization method in Section \ref{subsec:hetero}.} 
% In heterogeneous quantization, most previous works \cite{fromm2018heterogeneous,dong2019hawq} focus on the latter approach. The exception is \cite{wang2019haq} based on neural architecture search using reinforcement learning, but they could hardly achieve extremely low bit-width (i.e. less than $4$-bit). Meanwhile, we can effectively learn bit-widths using DropBits and validate low-bit quantized winning tickets in Section \ref{subsec:hetero}.

% \vspace{-0.1cm}

\section{Experiments}\label{sec:exp}

\begin{table}[t]
    \vskip -6pt
	\centering
	{
	\caption{Top-1/Top-5 error (\%) with ResNet-18 and MobileNetV2 on the ImageNet dataset.}
	\footnotesize
	\begin{tabular}{l| c| cc}
		\toprule
		Method & \makecell{\# Bits \\ W./A.} & \makecell{ResNet-18 \\ Top-1/Top-5} & \makecell{MobileNetV2 \\ Top-1/Top-5} \\
		\midrule
		Full-precision & $32/32$ & $30.24 / 10.92$ & $28.12/9.71$\\
		\midrule
		RQ \cite{louizos2018relaxed} & $4/4$ & $38.48 / 16.01$ & $- / -$  \\
		\midrule
		RQ ST \cite{louizos2018relaxed} & $4/4$ & $37.54$ / $15.22$ & $-$ / $-$ \\
% 		RQ ST \cite{louizos2018relaxed} & $4/4$ & $37.54 / 15.22$ & $-/-$ \\
        \midrule
        QIL\footnotemark[3] \cite{jung2019learning} & $4/4$ & $31.05 / 11.23$ & $32.77 / 12.51$ \\ 
		\midrule
		LLSQF \cite{zhao2020linear} & $4/4$ & $30.60 / 11.28$ & $32.63 / 12.01$ \\
		& $3/3$ & $33.33 / 12.58$ & $- / -$ \\
		\midrule
        TQT \cite{jain2019tqt, uhlich2020mixed} & $4/4$ & $30.49 / -$ & $32.21 / -$ \\
		\midrule
		\textbf{CPQ +} & $4 / 4$ & $\mathbf{30.37} / \mathbf{10.96}$ & $\mathbf{30.83} / \mathbf{11.26}$ \\
		\textbf{DropBits} & $3 / 3$ & $\mathbf{32.79}/\mathbf{12.57}$ & $\mathbf{35.71} / \mathbf{14.36}$ \\
		\bottomrule
	\end{tabular}
	\vskip -5pt
	\label{tab:imagenet}
	}
\end{table}
\addtocounter{footnote}{3}
\footnotetext[3]{Our own implementation with all layers quantized by using pretrained models available from PyTorch}

% \vspace{-0.2cm}
As popular deep learning libraries such as TensorFlow \cite{abadi2016tensorflow} and PyTorch from v1.3 \cite{paszke2019pytorch} already provide their own 8-bit quantization functionalities, we focus on low bit-width regimes ($2\!\sim\!4$-bit). In contrast to some other quantization papers, our method uniformly quantizes weights and activations of \emph{all} layers %in a layer-wise fashion, 
containing both the \emph{first} and \emph{last} layers. 
We first show that CPQ and DropBits have its own contribution, none of which is negligible. Then, we evaluate CPQ + DropBits on a large-scale dataset with deep networks. %(the results of experiments on MNIST and CIFAR-10 are deferred to Appendix). 
Finally, we demonstrate our heterogeneous quantization method yields promising results even if all layers have at most $4$-bit and validate a new hypothesis for quantization in Section \ref{subsec:lottery}.

\begin{table*}[t]
    \vskip -6pt
    \centering
	{
	\caption{Test error (\%) for quantized sub-networks using LeNet-5 on MNIST, VGG-7 on CIFAR-10,and ResNet-18 on ImageNet. Here, an underline means the learned bit-width and  ``T'' stands for ternary precision.}
	\footnotesize
	\begin{tabular}{c|cc|ccc}
		\toprule
		Model & \makecell{Initial \\ \# Bits \\ W/A} & \makecell{Test \\ Error} & \makecell{Trained \\ W. Bits \\ per layer} & \makecell{Test \\ Error \\ (Fixed)}  & \makecell{Test \\ Error \\ (Reg.)} \\
		\midrule
		& $4/4$ & $0.53$  & $4/4/\underline{3}/4$ & $0.55$ & $\mathbf{0.52}$ \\
        % & & & $4/\underline{3}/\underline{3}/4$ & $0.59$ & $\mathbf{0.58}$ \\[0.1cm]
        % \cline{2-6}
        LeNet-5 & $3/3$ & $0.58$ & $3/\underline{2}/3/3$ & $0.65$ & $\mathbf{0.55}$ \\
        % & & & $3/\underline{\text{T}}/3/\underline{2}$ & $0.85$  & $\mathbf{0.60}$ \\[0.1cm]
        % \cline{2-6}
        & $2/2$ & $0.63$ & $2/2/2/\underline{\text{T}}$ & $0.68$ & $\mathbf{0.59}$ \\
        % & & & $\underline{\text{T}}/2/2/\underline{\text{T}}$ & $0.70$ & $\mathbf{0.64}$ \\
        \midrule
		& $4/4$ & $6.77$ & \makecell{$4/4/4/4/4/\underline{3}/\underline{3}/4$} & $6.74$ & $\mathbf{6.65}$ \\
% 		& & & \makecell{$4/\underline{3}/4/4/4/\underline{3}/\underline{3}/4$} & $6.87$ & $\mathbf{6.80}$ \\[0.05cm]
% 		\cline{2-6}
		VGG-7 & $3/3$ & $6.82$ & \makecell{$3/3/3/3/3/\underline{2}/3/3$} & $6.81$ & $\mathbf{6.77}$ \\
% 		& & & \makecell{$3/\underline{2}/\underline{2}/\underline{2}/\underline{2}/\underline{\text{T}}/\underline{2}/3$} & $7.13$ & $\mathbf{7.04}$ \\[0.05cm]
% 		\cline{2-6}
		& $2/2$ & $7.49$ & \makecell{$2/2/2/2/2/2/2/\underline{\text{T}}$} & $7.43$ & $\mathbf{7.36}$ \\
% 		& & & \makecell{$\underline{\text{T}}/\underline{\text{T}}/\underline{\text{T}}/\underline{\text{T}}/\underline{\text{T}}/\underline{\text{T}}/2/\underline{\text{T}}$} & $9.62$ & $\mathbf{7.55}$ \\
		\midrule
		ResNet-18 & \makecell{$4/4$ \\ $3/3$} & \makecell{$33.20$ \\ $37.80$} & \makecell{$4/\underline{3}/\underline{3}/\underline{3}/\underline{3}/\underline{3}/\underline{3}/\underline{3}/\underline{3}/\underline{3}/\underline{3}/\underline{3}/\underline{3}/\underline{3}/\underline{3}/4/4/\underline{3}/4/4/4$ \\ $3/3/\underline{2}/3/\underline{2}/3/3/3/3/3/3/3/\underline{2}/3/3/3/3/3/3/3/3$} & \makecell{$34.58$ \\ $41.01$} & \makecell{$\mathbf{34.30}$ \\ $\mathbf{40.30}$} \\ % ~\\
% 		ResNet-18 & & & \makecell{$\underline{3}/\underline{3}/\underline{3}/\underline{3}/\underline{3}/\underline{3}/\underline{3}/\underline{3}/\underline{3}/\underline{3}/\underline{3}/\underline{3}/\underline{3}/\underline{3}/\underline{3}/\underline{3}/\underline{3}/\underline{3}/\underline{3}/\underline{3}/4$} & $36.46$ & $\mathbf{34.94}$ \\[0.05cm] % ~\\
% 		\cline{2-6}
% 		& $3/3$ & $37.80$ & \makecell{$3/3/\underline{2}/3/\underline{2}/3/3/3/3/3/3/3/\underline{2}/3/3/3/3/3/3/3/3$} & $41.01$ & $\mathbf{40.30}$ \\ % ~\\
% 		& & & \makecell{$3/3/\underline{2}/\underline{2}/\underline{2}/\underline{2}/3/3/3/3/3/3/\underline{2}/3/3/3/3/3/3/3/3$} & $43.41$ & $\mathbf{42.13}$ \\
% 			\midrule
% 			ResNet-34 & $4/4$ & $30.05$ & \makecell{all $3$-bit layers except $4$-bit $\{\text{first}, 28, 30, 31, 33, 34, 35, 36, \text{last}\}$ layers} & $32.33$ & $\mathbf{31.12}$ \\[0.05cm] % ~\\
% 			& & &
% 			\makecell{all $3$-bit layers except $4$-bit last layer} & $34.37$ & $\mathbf{31.47}$ \\ % ~\\
		\bottomrule
	\end{tabular}
	\vskip -3pt
% 		\vspace{-0.1cm}
	\label{tab:heterogeneous}
	}
\end{table*}

% \vspace{-0.1cm}
\subsection{Ablation Studies}\label{subsec:ablation}
To validate the efficacy of CPQ and DropBits, we successively apply each piece of our method to LeNet-5 \cite{lecun1998gradient} on MNIST and VGG-7 \cite{simonyan2014very} on CIFAR-10. Table \ref{tab:ablation} shows that CPQ outperforms RQ in most cases. One might wonder that the performance of RQ can be improved by an annealing schedule of the temperature in the Gumbel-Softmax trick. 
Unfortunately, RQ with an annealing schedule suffers from high variance of gradients due to low temperatures at the end of training as shown in Figure \ref{fig:learning_curves}, thus giving rise to worse performance than RQ as shown in Table \ref{tab:ablation}. % As a result, annealing the temperature in the Gumbel-Softmax trick gives rise to worse performance than RQ as shown in Table \ref{tab:ablation}.
% , the performance of RQ with annealing $\tau$ is rather poor than that of RQ. %RQ with an annealing schedule cannot beat RQ as seen in Table \ref{tab:ablation}. 
% However, CPQ does not suffer from the variance issue at all, thereby displaying the best learning curve in Figure \ref{fig:learning_curves}.
% Figure \ref{fig:learning_curves} also shows that the learning curve of RQ with an annealing schedule is below that of RQ at the end of training due to high variance of gradients arising from low temperatures.
Finally, it can be clearly identified that DropBits consistently improves CPQ by decreasing the bias of our multi-class STE in CPQ.

\subsection{ResNet-18 and MobileNetV2 on ImageNet}\label{subsec:imagenet}

To verify the effectiveness of our algorithm on the ImageNet dataset, we quantize the ResNet-18 \cite{he2016deep} and MobileNetV2 \cite{sandler2018mobilenetv2} architectures initialized with each pre-trained full-precision network available from the official PyTorch repository.
% in quite a different setting from \cite{louizos2018relaxed}. 
In Table \ref{tab:imagenet}, our method is only compared to the state-of-the-art algorithms that quantize both weights and activations of \emph{all} layers for fair comparisons. The extensive comparison against recent works that remain the first or last layer in the full-precision is given in Appendix.

Table \ref{tab:imagenet} illustrates how much better our model performs than the latest quantization methods. 
In ResNet-18, CPQ + DropBits outdoes RQ, QIL, LLSQF, and TQT, even achieving the top-1 and top-5 errors in $4$-bit nearly close to those of the full-precision network. 
In MobileNetV2, CPQ + DropBits with 4-bit surpasses all existing studies by more than one percentage point. Moreover, we quantize MobileNetV2 to $3$-bit, obtaining competitive performance, which is remarkable due to the fact that none of previous works successfully quantizes MobileNetV2 to $3$-bit.

% \vspace{-0.1cm}
\subsection{Finding Quantized Sub-networks}\label{subsec:hetero}
In this experiment, we validate a \emph{new hypothesis for quantization} by training the probabilities of binary masks using the regularizer in Section \ref{subsec:learning_bit} to learn the bit-width of each layer.
For brevity, only weights are heterogeneously quantized, and the bit-width for activations remains fixed.

% \vspace{-0.1cm}
In Table \ref{tab:heterogeneous}, the fourth column represents the bit-width per layer learned by our regularizer, and the fifth and last columns indicate the test error when fixing the bit-width of each layer same as trained bit-widths (fourth column) from scratch and when using our regularization approach, respectively. Table \ref{tab:heterogeneous} shows that a learned structure by our heterogeneous quantization method (last column) is superior to the fixed structure with learned bit-widths from scratch (fifth column) for all cases. It might be doubtful whether our regularizer is able to recognize which layer is really redundant or not. This may be indirectly substantiated by the observation that the fixed structure with trained bit-widths from scratch (fifth column) outperforms the uniform quantization (third column) on CIFAR-10. More experiments on different values of the regularization parameter $\lambda$ are deferred to Appendix.

% \vspace{-0.1cm}
\section{Conclusion}\label{sec:conclusion}
% \vspace{-0.2cm}
We proposed \emph{Cluster-Promoting Quantization (CPQ)}, which not only finds the optimal quantization grids but also encourages the underlying full-precision weights to cluster around those quantization grids cohesively in low bit-width regimes. To reduce the bias of the multi-class STE in CPQ, we also proposed a novel bit-drop technique, \emph{DropBits}. We showed that both CPQ and DropBits possess its own value, thereby leading CPQ + DropBits to achieve the state-of-the-art performance for ResNet-18 and MobileNetV2 on ImageNet. Furthermore, we took one step forward to consider heterogeneous quantization by simply penalizing binary masks in DropBits, which enables us to find out quantized sub-networks. As future work, we plan to extend our heterogeneous quantization method to activations and its application to other quantizers.
\vspace{-0.2cm}

\section*{Acknowledgements}

\vspace{-0.1cm}
This work was supported by the National Research Foundation of Korea (NRF) grants (2018R1A5A1059921, 2019R1C1C1009192) and Institute of Information \& Communications Technology Planning \& Evaluation (IITP) grants (No. 2017-0-01779, A machine learning and statistical inference framework for explainable artificial intelligence, and No.2019-0-00075, Artificial Intelligence Graduate School Program(KAIST)) funded by the Korea government(MSIT). 
% This work was partly supported by Samsung Research Funding \& Incubation Center of Samsung Electronics under Project Number SRFC-IT1702-15.

{\small
\bibliographystyle{ieee_fullname}
\bibliography{reference}
}

\newpage
\onecolumn
\appendix
% \section*{Supplementary Materials}
\section{Algorithm of Cluster-Promoting Quantization with DropBits}

\begin{algorithm}[H]
 	\caption{\textbf{C}luster-\textbf{P}romoting \textbf{Q}uantization (CPQ) + DropBits}
 	\label{alg:srq_training}
 	\begin{algorithmic}[1]
 		\State {\bfseries Input:} Training data $\mathcal{D}$, network parameters $\{W_l, b_l\}_{l=1}^{L}$, layer-wise grid interval parameters and the standard deviations of a logistic distribution in the $l$-th layer $\{\alpha_l, \sigma_l\}_{l=1}^{L}$. The probability parameters $\{\Pi_l^{(k)}\}_{(l,k)=(1,0)}^{(L,b-1)}$ for DropBits. $\Pi_l^{(k)}$ is the probability parameter for $k$-th DropBits mask at the $l$-th layer. 
 		\State {\bfseries Output:} A low bit-width model with quantized network parameters $\{\widehat W_l, \widehat b_l\}_{l=1}^{L}$ after deployment procedure.
 		\State {\bfseries Initialize:} A bit-width $b$ and parameters $\{W_l, b_l, \alpha_l, \sigma_l\}_{l=1}^{L}$. Initialize layer-wise grid $\widehat{\mathcal{G}}_l \coloneqq [g_{l,0}, g_{l,1}, \cdots, g_{l, 2^b-1}] = \alpha_l [-2^{b-1}, \cdots, 2^{b-1} - 1]$ for $l \in \{1, \cdots, L\}$. 
        \Procedure{Training}{}
        \State // ~Forward pass
        \For{$l = 1, \cdots, L$}
            \Let {$x$}{Each entry of $W_l$ or $b_l$}
            \State {$I_l = \widehat{\mathcal{G}}_l - \alpha/2$}
            \Comment{Shift the grid by $-\alpha/2$}
    		\State {$F = \text{Sigmoid}\Big(\frac{I_l - x}{\sigma_l}\Big)$} \Comment{Compute CDFs}
    		\State {$\pi_i = {F[i+1] - F[i]} \,\, \text{for } i = 0, \cdots, 2^b-1$} \Comment{Eq. \eqref{eqn:unnorm_prob}}
            %\If{DropBits}
            \State // DropBits (Line $12 \sim 14$)
            \State {Sample a mask $Z_k$ with probability $\Pi_k$ for each $k = 0, \cdots, b-1$} \Comment{Eq. \eqref{eq:mask}} 
            \State $\bar{\pi} = \pi \odot Z$ \Comment{Figure \ref{fig:mask_design}}
            \State {$\widetilde{\pi}_i = \bar{\pi}_i/\sum_{j=0}^{2^b - 1} \bar{\pi}_j$} \Comment{Sum-to-1 Normalization}
    		%\Else
    		%\State {$\widetilde{\pi} = \pi$}
    		%\EndIf
    		\State $y = \mathtt{one\_hot}[\argmax_i \widetilde{\pi}_i]$ \Comment{Eq. \eqref{eq:one_hot}}
    		\State $\widehat x = y \odot \widehat{\mathcal{G}_l}$ \Comment{Quantization}
    		\State Activation can be quantized in the same way, but we do not apply DropBits to activations
        \EndFor
        \State // ~Backward pass
        \For{$l = L, \cdots, 1$}
            \State Compute gradients $(\frac{\partial \mathcal{L}}{\partial W_l}, \frac{\partial \mathcal{L}}{\partial b_l}, \frac{\partial \mathcal{L}}{\partial \alpha_l}, \frac{\partial \mathcal{L}}{\partial \sigma_l}, \frac{\partial\mathcal{L}}{\partial \Pi_l^{(k)}})$ \Comment{Eq. \eqref{eq:ste}}
            \State Update parameters $(W_l, b_l, \alpha_l, \sigma_l, \Pi_l^{(k)})$
        \EndFor
		\EndProcedure
		\Procedure{Deployment}{}
		    \For{$l = 1, \cdots, L$}
		        \State $\widehat W_l = \min(\max(\alpha_l \cdot \mathrm{Round}(W_l/\alpha_l), g_{l,0}), g_{l,2^b-1})$
		        \State $\widehat b_l = \min(\max(\alpha_l \cdot \mathrm{Round}(b_l/\alpha_l), g_{l,0}), g_{l,2^b-1})$
		    \EndFor
		\EndProcedure
 	\end{algorithmic}
\end{algorithm}

\newpage
\section{Extensive Comparison for ResNet-18 and MobileNetV2 on ImageNet}
Our method, CPQ + DropBits surpasses quantization methods remaining the first or last layer in the full precision as well as the latest algorithms that quantize both the weights and activations of all layers including the first and last layers, except QIL \cite{jung2019learning} and LSQ \cite{esser2020learned} both of which utilize the full-precision first and last layer as well as employ their own ResNet-18 pretrained model performing much higher than one available from the official PyTorch repository.

\begin{table}[H]
	\vspace{-0.2cm}
    \centering
	{\footnotesize
	\caption{Top-1/Top-5 error (\%) with ResNet-18 and MobileNetV2 on ImageNet using 4-bit. $\dagger$ denotes the use of the full-precision first or last layer, and $\ddagger$ indicates our own implementation with all layers quantized by  using pretrained models available from the official PyTorch repository.}
	\setlength{\extrarowheight}{0.05cm}
	\setlength\tabcolsep{8pt}
% 	\vspace{+0.1cm}
	\begin{tabular}{l|cc|cc}
		\toprule
		& \multicolumn{2}{c}{ResNet-18} & \multicolumn{2}{c}{MobileNetV2} \\
		\midrule
		Methods & Top-$1$ & Top-$5$ & Top-$1$ & Top-$5$ \\
		\midrule
		Full-precision & $30.24$ & $10.92$ & $28.12$ & $9.71$ \\
		\midrule
		DoReFa\textsuperscript{\textdagger} \cite{Zhou2016dorefa} & $31.9$ & $-$ & $-$ & $-$ \\
		BCGD\textsuperscript{\textdagger} \cite{yin2018bcgd} & $32.64$ & $12.24$ & $-$ & $-$ \\
		LQ-Nets\textsuperscript{\textdagger} \cite{zhang2018lqnets} & $30.7$ & $11.2$ & $-$ & $-$ \\
		PACT\textsuperscript{\textdagger} \cite{choi2018pact,wang2019haq} & $30.8$ & $-$ & $38.56$ & $17.30$ \\
		RQ \cite{louizos2018relaxed} & $38.48$ & $16.01$ & $-$ & $-$ \\
		RQ ST \cite{louizos2018relaxed} & $37.54$ & $15.22$ & $-$ & $-$ \\
		DSQ\textsuperscript{\textdagger} \cite{gong2019dsq} & $30.44$ & - & $35.20$ & $-$ \\
		QIL\textsuperscript{\textdaggerdbl} \cite{jung2019learning} & $31.05$ & $11.23$ & $32.77$ & $12.51$ \\
		QIL\textsuperscript{\textdagger} \cite{jung2019learning} & $29.90$ & $-$ & $-$ & $-$ \\
		TQT \cite{jain2019tqt, uhlich2020mixed} & $30.49$ & $-$ & $32.21$ & $-$\\
		LLSQF \cite{zhao2020linear} & $30.60$ & $11.28$ & $32.63$ & $12.01$ \\
		LSQ\textsuperscript{\textdagger} \cite{esser2020learned} & $28.90$ & $10.0$ & $-$ & $-$ \\
		\midrule
		\textbf{CPQ + DropBits (Ours)} & $\mathbf{30.37}$ & $\mathbf{10.96}$ & $\mathbf{30.83}$ & $\mathbf{11.26}$ \\ 
		\bottomrule
	\end{tabular}
	\vskip -10pt
% 	\vspace{-0.1cm}
	\label{tab:extensive}
	}
\end{table}

\newpage
\section{More Experiments on Heterogeneous Quantization}\label{sec:hetero_total}
As we can see in Table \ref{tab:heterogeneous_total}, our heterogeneous quantization method is capable of finding quantized sub-networks in a broad range of regularization parameter $\lambda$.

\begin{table}[H]
    \centering
	{
	\caption{Test error (\%) for quantized sub-networks using LeNet-5 on MNIST, VGG-7 on CIFAR-10,and ResNet-18 on ImageNet. Here, an underline means the learned bit-width and  ``T'' stands for ternary precision.}
	%\setlength\tabcolsep{2pt}
% 	\resizebox{\linewidth}{!}{
	\footnotesize
	\begin{tabular}{c|cc|ccc}
		\toprule
		Model & \makecell{Initial \\ \# Bits \\ W/A} & \makecell{Test \\ Error} & \makecell{Trained \\ W. Bits \\ per layer} & \makecell{Test \\ Error \\ (Fixed)}  & \makecell{Test \\ Error \\ (Reg.)} \\
		\midrule
		& $4/4$ & $0.53$  & $4/4/\underline{3}/4$ & $0.55$ & $\mathbf{0.52}$ \\
        & & & $4/\underline{3}/\underline{3}/4$ & $0.59$ & $\mathbf{0.58}$ \\[0.1cm]
        \cline{2-6}
        LeNet-5 & $3/3$ & $0.58$ & $3/\underline{2}/3/3$ & $0.65$ & $\mathbf{0.55}$ \\
        & & & $3/\underline{\text{T}}/3/\underline{2}$ & $0.85$  & $\mathbf{0.60}$ \\[0.1cm]
        \cline{2-6}
        & $2/2$ & $0.63$ & $2/2/2/\underline{\text{T}}$ & $0.68$ & $\mathbf{0.59}$ \\
        & & & $\underline{\text{T}}/2/2/\underline{\text{T}}$ & $0.70$ & $\mathbf{0.64}$ \\
        \midrule
		& $4/4$ & $6.77$ & \makecell{$4/4/4/4/4/\underline{3}/\underline{3}/4$} & $6.74$ & $\mathbf{6.65}$ \\
		& & & \makecell{$4/\underline{3}/4/4/4/\underline{3}/\underline{3}/4$} & $6.87$ & $\mathbf{6.80}$ \\[0.1cm]
		\cline{2-6}
		VGG-7 & $3/3$ & $6.82$ & \makecell{$3/3/3/3/3/\underline{2}/3/3$} & $6.81$ & $\mathbf{6.77}$ \\
		& & & \makecell{$3/\underline{2}/\underline{2}/\underline{2}/\underline{2}/\underline{\text{T}}/\underline{2}/3$} & $7.13$ & $\mathbf{7.04}$ \\[0.1cm]
		\cline{2-6}
		& $2/2$ & $7.49$ & \makecell{$2/2/2/2/2/2/2/\underline{\text{T}}$} & $7.43$ & $\mathbf{7.36}$ \\
		& & & \makecell{$\underline{\text{T}}/\underline{\text{T}}/\underline{\text{T}}/\underline{\text{T}}/\underline{\text{T}}/\underline{\text{T}}/2/\underline{\text{T}}$} & $9.62$ & $\mathbf{7.55}$ \\
		\midrule
		& $4/4$ & $33.20$ & $4/\underline{3}/\underline{3}/\underline{3}/\underline{3}/\underline{3}/\underline{3}/\underline{3}/\underline{3}/\underline{3}/\underline{3}/\underline{3}/\underline{3}/\underline{3}/\underline{3}/4/4/\underline{3}/4/4/4$ & $34.58$ & $\mathbf{34.30}$  \\ 
		ResNet-18 & & & \makecell{$\underline{3}/\underline{3}/\underline{3}/\underline{3}/\underline{3}/\underline{3}/\underline{3}/\underline{3}/\underline{3}/\underline{3}/\underline{3}/\underline{3}/\underline{3}/\underline{3}/\underline{3}/\underline{3}/\underline{3}/\underline{3}/\underline{3}/\underline{3}/4$} & $36.46$ & $\mathbf{34.94}$ \\[0.1cm] % ~\\
		\cline{2-6}
		& $3/3$ & $37.80$ & \makecell{$3/3/\underline{2}/3/\underline{2}/3/3/3/3/3/3/3/\underline{2}/3/3/3/3/3/3/3/3$} & $41.01$ & $\mathbf{40.30}$ \\ % ~\\
		& & & \makecell{$3/3/\underline{2}/\underline{2}/\underline{2}/\underline{2}/3/3/3/3/3/3/\underline{2}/3/3/3/3/3/3/3/3$} & $43.41$ & $\mathbf{42.13}$ \\
		\bottomrule
	\end{tabular}
% 	}
	\vskip -10pt
% 		\vspace{-0.1cm}
	\label{tab:heterogeneous_total}
	}
\end{table}

\newpage
\section{Proof of Proposition \ref{prop}}
\begin{align*}
    {\partial \mathcal{L} \over \partial x} = \Sigma_i{\partial \mathcal{L} \over \partial \pi_{i}}{\partial \pi_{i} \over \partial x} = {\partial \mathcal{L} \over \partial y_{i_{\text{max}}}}{\partial \pi_{i_{\text{max}}} \over \partial x} \,\, (\because \eqref{eq:ste})    
\end{align*}
where
\begin{align*}
    {\partial\pi_{i_{\text{max}}} \over \partial x} =
    &{1 \over \sigma}\big(\mathrm{Sigmoid}\big( {g_{i_{\text{max}}} + \frac{\alpha}{2} - x \over \sigma} \big)\mathrm{Sigmoid}\big(-{g_{i_{\text{max}}} + \frac{\alpha}{2} - x \over \sigma}\big) \\ 
    &- \mathrm{Sigmoid}\big( {g_{i_{\text{max}}} - \frac{\alpha}{2} - x \over \sigma}\big)\mathrm{Sigmoid}\big(- {g_{i_{\text{max}}} - \frac{\alpha}{2} - x \over \sigma} \big)\big).
\end{align*}
Let $\lvert{\partial \mathcal{L} \over \partial y_{i_{\text{max}}}}\rvert$ be bounded by $M$. Since ${\partial\pi_{i_{\text{max}}} \over \partial x}\rvert_{x = g_{i_{\text{max}}}} = 0$ and ${\partial\pi_{i_{\text{max}}} \over \partial x}$ is continuous at $g_{i_{\text{max}}}$ and symmetric about $(g_{i_{\text{max}}}, 0)$, for any $\epsilon > 0$, there exists $\delta > 0$ such that $\lvert{\partial\pi_{i_{\text{max}}} \over \partial x}\rvert$ is bounded by $\epsilon/\max(M, 1)$ for every $x \in (g_{i_{\text{max}}}-\delta, g_{i_{\text{max}}}+\delta)$ so that $\lvert{\partial \mathcal{L} \over \partial x}\rvert$ is bounded by $\epsilon$. Hence, ${\partial \mathcal{L} \over \partial x}$ converges to zero as $\delta$ goes to zero, which implies that $x$ approaches $g_{i_{\text{max}}}$. % that is inversely proportional to $\lvert{\partial \mathcal{L} \over \partial y_{i_{\text{max}}}}\rvert$ % Note that ${\partial \mathcal{L} \over \partial x}$ is exactly zero when $x$ is located on $g_{i_{\text{max}}}$ since ${\partial\pi_{i_{\text{max}}} \over \partial x}\rvert_{x = g_{i_{\text{max}}}} = 0$.

\newpage
\section{Comparison of CPQ + DropBits with Gumbel-Softmax + multi-class STE}

To show the effectiveness of CPQ + DropBits further, we empirically compare it with an algorithm using the Gumbel-Softmax STE in the forward pass instead of DropBits and our multi-class STE in the backwared pass. Let such an algorithm be called “Gumbel-Softmax + multi-class STE”.

\begin{table}[H]
% 	\vskip -15pt
    \caption{Test error ($\%$) for LeNet-5 on MNIST and VGG-7 on CIFAR-10.}
	\centering
	{\footnotesize
	\setlength\tabcolsep{2pt}
	\resizebox{6.8cm}{!}{
	\begin{tabular}{c|c|c|c|c}
	\toprule
	Dataset & RQ & \makecell{\# Bits \\ W./A.} & \makecell{Gumbel-Softmax \\ + multi-class STE}& \makecell{CPQ + \\DropBits} \\
	\midrule
	& $4/4$ & $0.58$ & $0.57$ & $\mathbf{0.53}$ \\
	MNIST & $3/3$ & $0.69$ & $0.65$ & $\mathbf{0.58}$ \\
	& $2/2$ & $0.76$ & $0.74$ & $\mathbf{0.63}$ \\
	\midrule
	& $4/4$ & $8.43$ & $8.25$ & $\mathbf{6.85}$ \\
	CIFAR-10 & $3/3$ & $9.56$ & $8.44$ & $\mathbf{6.94}$ \\
	& $2/2$ & $11.75$ & $10.41$ & $\mathbf{7.51}$ \\
	\bottomrule 
	\end{tabular}
	}
	}
	\label{tab:reviwer1}
\end{table}

Thanks to our multi-class STE in the backward pass, Gumbel-Softmax + multi-class STE performs better than RQ, but still worse than CPQ + DropBits. This seems primarily due to the fact that Gumbel-Softmax still incurs large quantization error after training.

\newpage
\section{Implementation Details}\label{sec:exp_detail}
% Basically, we follow the same experimental settings as introduced in Appendix A in \cite{louizos2018relaxed} for fair comparison. 
The weights and activations of all layers including the first and last layers (denoted by $W$ and $A$) are assumed to be perturbed as $\widetilde{W} = W + \epsilon$ and $\widetilde{A} = A + \epsilon$ respectively, under $\epsilon \sim L(0, \sigma)$ as we describe in Section 2. 

Concerning DropBits regularization in \ref{subsec:dropbits}, we initialize the probability of each binary mask with $\Pi \sim  \mathcal{N}(0.9, 0.01^2)$ (i.e. corresponding to low dropout probability). % since the high dropout rate could make training a network unstable.
The concrete distribution of a binary mask is stretched to $\zeta=1.1$ and $\gamma=-0.1$ as recommended in \cite{louizos2017l0reg}, and $\tau'$ is initialized to $0.2$ to make a binary mask more discretized. 

For MNIST experiments, we train LeNet-5 with 32C5 - MP2 - 64C5 - MP2 - 512FC - Softmax architecture for 100 epochs irrespective of the bit-width. In addition, a learning rate is set to 5e-4 regardless of the bit-width and exponentially decayed with decay factor $0.8$ for the last 50 epochs. The input is normalized into $[-1, 1]$ range without any data augmentation.

For CIFAR-10 experiments, following the convention that the location of max-pooling layer is changed, which originates from \cite{Rastegari2016xnornet}, a max-pooling layer is located after a convolutional layer, but before a batch normalization and an activation function. We train VGG-7 with 2x(128C3) - MP2 - 2x(256C3) - MP2 - 2x(512C3) - MP2 - 1024FC - Softmax architecture for 300 epochs, and a learning rate is initially set to 1e-4 regardless of the bit-width. The learning rate is multiplied by $0.1$ at $50\%$ of the total epochs and decay exponentially with the decay factor 0.9 during the last 50 epochs. The input images are preprocessed by substracting its mean and dividing by its standard deviation. The training set is augmented as follows: (i) a random $32\times32$ crop is sampled from a padded image with 4 pixels on each side, (ii) images are randomly flipped horizontally. The test set is evaluated without any padding or cropping. Note that a batch normalization layer is put after every convolutional layer in VGG-7, but not in LeNet-5.

In Section \ref{subsec:ablation}, RQ with an annealing schedule of the temperature $\tau$ in RQ is implemented by following \cite{Jang2016gumbel}: $\tau$ is annealed every $1000$ iterations by the schedule $\tau = \max(0.5, \exp{(-t/100000)})$ in $3$-bit and $\tau = \max(0.5, 2\exp{(-t/100000)})$ in $4$-bit in order to make the decreasing rate of $\tau$ as small as possible. Here, $t$ is the global training iteration.

For ImageNet experiments in Section \ref{subsec:imagenet}, the weight parameters of both ResNet-18 and MobileNetv2 are initialized with the pre-trained full precision model available from the official PyTorch repository. When quantizing ResNet-18 to $3$-bit, fine-tuning is implemented for $80$ epochs with a batch size of $256$: a learning rate is initialized to 2e-5 and divided by two at $50$, $60$, and $68$ epochs. When ResNet-18 is quantized to $4$-bit, fine-tuning is carried out for $150$ epochs with a batch size of $128$: for the first $125$ epochs, a learning rate is set to 5e-6, but 1e-6 for the last $25$ epochs. When quantizing MobileNetV2 to $3$-bit, fine-tuning is done for $140$ epochs with a batch size of $96$: an initial learning rate is initialized to 2e-5 and divided by two at $10$ and $20$. When MobileNetV2 is quantized to $4$-bit, fine-tuning is performed for $25$ epochs with a batch size of $48$ and an initial learning rate of 2e-5: the learning rate is divided by two at $10$, $12$, $18$, and $20$ epochs for $4$-bit. We employ AdamW in Decoupled Weight Decay Regularization \cite{loshchilov2018decoupled} with a weight decay factor of $0.01$.

In Section \ref{subsec:hetero} and \ref{sec:hetero_total}, if the probability of a binary mask is less than 0.5, then we drop the corresponding bits. For LeNet-5 on MNIST and VGG-7 on CIFAR-10, our regularization term in Section \ref{subsec:learning_bit} is activated only for the first $50\%$ of the total epochs. With the remained bit-width for each layer, fine-tuning process is conducted for the last $50\%$ of the total epochs. For ResNet-18 on ImageNet, we initialize the weights of ResNet-18 with the pre-trained full precision model and train it for ten epochs for simplicity. During training, our regularization term in Section \ref{subsec:learning_bit} is activated only for the first $9$ epochs, and fine-tuning process is done for the last epoch with the remained bit-width of each layer fixed. All experiments in Table \ref{tab:heterogeneous} and \ref{tab:heterogeneous_total} were conducted by the use of AdamW: the weight decay value is set to $0.01$ for LeNet-5, $0.02$ for VGG-7, and $0.01$ for ResNet-18. We consider the regularization parameter $\lambda \in [5 \times 10^{-5}, 10^{-2}]$ to encourage layer-wise heterogeneity.

\end{document}